\documentclass{article}

\usepackage{microtype}
\usepackage{graphicx}
\usepackage{subfigure}
\usepackage{booktabs} %

\usepackage{mystyle}

\usepackage[accepted]{icml2023}

\usepackage{amsmath}
\usepackage{amssymb}
\usepackage{bbold}
\usepackage{mathtools}
\usepackage{amsthm}

\newcommand{\maxKSWD}{max K-SWD}

\begin{document}

\twocolumn[
\icmltitle{On the Convergence Rate of Gaussianization with Random Rotations}

\icmlsetsymbol{equal}{*}

\begin{icmlauthorlist}
\icmlauthor{Felix Draxler}{uhd}
\icmlauthor{Lars Kühmichel}{uhd}
\icmlauthor{Armand Rousselot}{uhd}
\icmlauthor{Jens Müller}{uhd}
\icmlauthor{Christoph Schnörr}{uhd}
\icmlauthor{Ullrich Köthe}{uhd}
\end{icmlauthorlist}

\icmlaffiliation{uhd}{Heidelberg University, Germany}

\icmlcorrespondingauthor{Felix Draxler}{felix.draxler@iwr.uni-heidelberg.de}

\icmlkeywords{Machine Learning, ICML}

\vskip 0.3in
]

\printAffiliationsAndNotice{}  %

\begin{abstract}
    Gaussianization \cite{chen_gaussianization_2000} is a simple generative model that can be trained without backpropagation. It has shown compelling performance on low dimensional data. As the dimension increases, however, it has been observed that the convergence speed slows down. We show analytically that the number of required layers scales linearly with the dimension for Gaussian input. We argue that this is because the model is unable to capture dependencies between dimensions. Empirically, we find the same linear increase in cost for arbitrary input $p(x)$, but observe favorable scaling for some distributions. We explore potential speed-ups and formulate challenges for further research.
\end{abstract}

\section{Introduction}

Generative modeling is one of the most active areas of research in Machine Learning. A plethora of different architectures based on neural networks have been proposed, including: generative adversarial networks \cite{goodfellow2020generative}, variational auto-encoders \cite{kingma2019introduction}, normalizing flows \cite{rezende_variational_2015}, and most recently denoising diffusion models \cite{ho2020denoising}.

Diffusion models currently take the lead in efficient training and high-quality sampling. However, evidence so far is mostly empirical and subject to change. This calls for a rigorous comparison of the different approaches.

For all models, universal approximation theorems guarantee that all reasonable distributions can be represented by the different models (e.g.~\citet{teshima_coupling-based_2020,teshima_universal_2020,koehler_representational_2021}). This is a strong statement in the sense that with enough resources, anything can be represented. However, these theorems are not useful for model selection, mainly as no statements about the required resources (model complexity, training speed, sample complexity) to achieve a specified performance are made. Also, existing results often limit themselves to relatively weak measures of convergence.

In this work, we consider a variant of normalizing flows called Gaussianization \cite{chen_gaussianization_2000} and its variants rotation-based iterative Gaussianization (RBIG) \cite{laparra_iterative_2011}, sliced iterative normalizing flows (SINF) \cite{dai_sliced_2021} and Gaussianization Flow (GF) \cite{meng_gaussianization_2020}. For the first time, we provide an explicit convergence rate for Gaussianization.

\begin{figure}[ht]
    \centering
    \includegraphics[width=.913\linewidth]{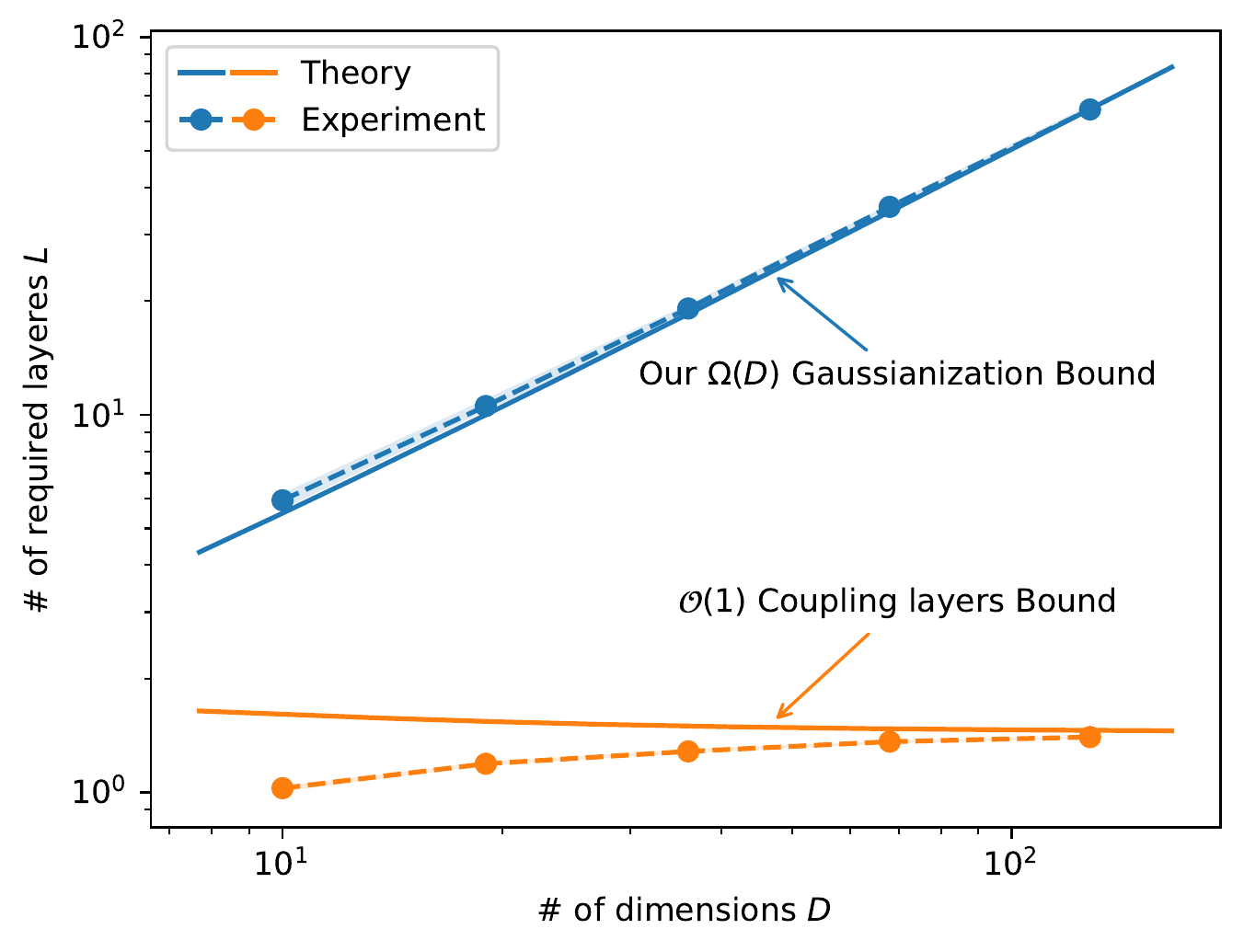}
    \caption{\textbf{Empirical scaling of learning Gaussian distributions as a function of dimension $D$ in the limit of low loss.} Gaussianization requires at least $\Omega(D)$ layers (\cref{sec:guarantee-random}), while only at most constant (i.e.~$\Oo(1)$) number of coupling layers are needed (\cref{sec:coupling-comparison}, based on \citet{draxler_whitening_2022}). The solid lines are the exact values predicted by the theories, the dots indicate experimental measurements. The shades show the inter-quartile range.}
    \label{fig:gaussian-scaling-experiment}
\end{figure}

In particular, we contribute:
\begin{itemize}
    \item We analytically derive that the number of Gaussianization layers required to achieve the same improvement in loss grows linearly with the dimensionality of the problem for Gaussian input and random rotations (see \cref{fig:gaussian-scaling-experiment,sec:guarantee-random}).
    \item We demonstrate limits of determining better-than-random rotations from finite training data (see \cref{sec:guarantee-learned-rotations}).
    \item We argue that this is due to the model being unable to capture dependencies between dimensions, which dominates in high dimensions (see \cref{sec:coupling-comparison}).
    \item We empirically determine the scaling behavior for real world datasets, where we find a similar linear increase in complexity with dimension, and favorable scaling for some distributions (see \cref{sec:dependency-real}).\footnote{Code available at: \url{https://github.com/vislearn/Gaussianization-Bound}}
\end{itemize}
Broadly speaking, we analyze the scaling behaviour of Gaussianization with dimension analytically and empirically. This draws a general picture over the convergence behavior of different invertible layers, see \cref{sec:Conclusion}.

\section{Related Work}

There is a large collection of different methods which iteratively transform between distributions via rotations and single-dimensional transforms. Originally, the idea of iteratively transporting input data to standard normal latent codes has been proposed by \citet{chen_gaussianization_2000}. \citet{laparra_iterative_2011} extended the idea with rotation-based iterative Gaussianization (RBIG), by also considering the reverse transport from latent codes to the data distribution. Other variants like Iterative Distribution Transfer (IDT) replace the standard normal by an arbitrary other distribution \cite{pitie_automated_2007}. \citet{meng_gaussianization_2020} leave the iterative scheme and train the flow end-to-end. They demonstrate that the resulting performs favorable in sitations of little data.

An important part in all these works is to find meaningful non-Gaussian projections of the data. Originally, random orthogonal matrices, ICA and PCA were suggested \cite{chen_gaussianization_2000}. %
\citet{meng_gaussianization_2020} learn rotations jointly with the dimension-wise transforms. Sliced Iterative Normalizing Flow (SINF) uses \maxKSWD, which optimizes for the $K$ most non-Gaussian directions in terms of Sliced Wasserstein Distance (SWD) \cite{dai_sliced_2021}. Sliced Wasserstein Flows (SWF) had previously suggested utilizing SWD, but used this to iteratively solve a PDE \cite{pmlr-v97-liutkus19a}.

There has been significant work on showing convergence guarantees for normalizing flows. Most work considers convergence under weak convergence or convergence under Wasserstein distance, both for Gaussianization \cite{chen_gaussianization_2000,meng_gaussianization_2020} and for coupling normalizing flows \cite{teshima_universal_2020,koehler_representational_2021}. While these measures of convergence ensure accurate samples ``$x \sim q(x) \to x \sim p(x)$'', %
they do not show convergence of the corresponding densities ``$q \to p$'' \cite{gibbs_choosing_2002}. Like \citet{draxler_whitening_2022} for coupling flows, we fill this gap by considering the KL divergence, a stronger notion of convergence. We also give explicit convergence rates instead of asymptotic guarantees. Our theoretical derivations are limited to Gaussian distributions close to convergence, however.

\section{Gaussianization Fundamentals}

\subsection{Model}
\label{sec:setup}

\begin{figure}
    \centering
    \includegraphics[width=.85\linewidth]{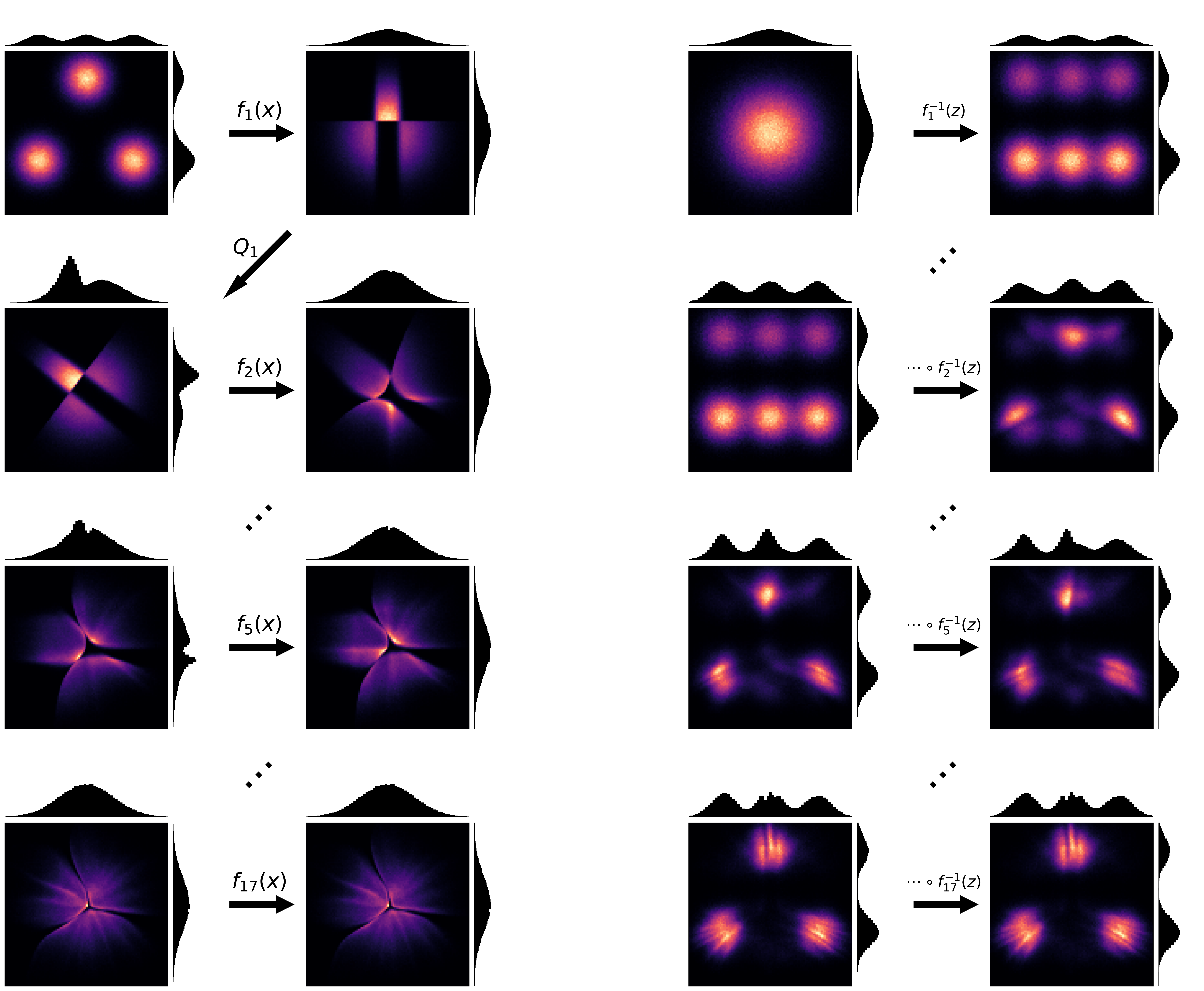}
    \caption{\textbf{Gaussianization learns a Gaussian mixture with three modes.} \textit{(Left)} Gaussianization makes marginals normal and rotates randomly. Iterating makes the latent distribution gradually Gaussian. \textit{(Right)} The first layer approximates $p(x)$ via the product of its marginals, see first row: $q_1(x) \approx p(x_1) p(x_2)$. Subsequent rows show the effect of additional later layers. The rows show the effect of layer $1, 2, 5$, and $17$.}
    \label{fig:iterative-gaussianization-visualization}
\end{figure}

Gaussianization is a variant of Normalizing Flows. Here, we follow the presentation in our previous work \citet{draxler_whitening_2022}. Flows represent a probability distribution via an invertible function $f_\theta(x)$ that maps samples $x$ from the unknown data distribution $p(x)$ to \textit{latent variables} $z = f_\theta(x)$ so that $z$ follow a simple distribution, typically the standard normal. The function $f_\theta$ then yields an estimate $q(x)$ for the true data distribution $p(x)$ via the change of variables formula (e.g.~\citet{rezende_variational_2015}):
\eql{
    q(x)
    = \Nn(f_\theta(x); 0, I) |\det J|,
}
where $J = \nabla f_\theta(x)$ is the Jacobian of $f_\theta(x)$. This is visualized in \cref{fig:iterative-gaussianization-visualization}.

A normalizing flow is often trained and evaluated using maximum likelihood, which is equivalent to minimizing the Kullback-Leibler divergence between the distribution of the latent code $q(z)$, as given by $z = f_\theta(x)$ when $x \sim p(x)$, and the standard normal:
\eqall{&
    \Ll 
    = \KL{q(z)}{\Nn(0, I)}  \\
    =&\,\, \EE_{x \sim p(x)}\left[\tfrac12 \norm{f_\theta(x)}^2 - \log |\det J|\right] + C.
}{\label{eq:loss}}
The additional term $C$ consists of the log partition sum of a standard normal distribution, and the unknown entropy of the data: $C = (D/2)\log(2\pi) - H[p(x)]$. It is independent of the model parameters and can therefore be dropped for optimization.

For sampling from $q(x)$, draw a latent code $z \in \Nn(0, I)$ and apply the inverse: $x = f_\theta^{-1}(z)$.

A useful invertible architecture $f_\theta$ has to (i)~be computationally easy to invert, (ii)~be able to represent complex transformations, and (iii)~have a tractable Jacobian determinant $|\det J|$ \cite{ardizzone_analyzing_2018}. A plethora of such architectures have been suggested, e.g.~\cite{kobyzev_normalizing_2021}.

In this work, we focus on Gaussianization, first presented by \cite{chen_gaussianization_2000}, and its variants rotation-based iterative Gaussianization (RBIG) \cite{laparra_iterative_2011}, sliced iterative normalizing flows (SINF) \cite{dai_sliced_2021} and Gaussianization Flow (GF) \cite{meng_gaussianization_2020}.
It is a deep architecture that consists of several blocks, each containing a rotation, and a \textit{dimension-wise} nonlinear transform:
\eql{
    \label{eq:gaussianization-block}
    f_\text{block}(x) = (f_\text{dim} \circ f_\text{rot})(x).
}

The dimension-wise nonlinear transform $f_\text{dim}$ treats each dimension $x_i, i = 1, \dots, D$ independently:
\eql{
    x_i' = f_{\text{dim}, \theta_i}(x_i),
}
where $\theta_i$ is a parameter vector defining the function for each dimension $i = 1, \dots, D$. Crucially, $x_i$ does not depend on $x_j$, for $i \neq j$.
Given this structure, ``sliced (iterative) normalizing flows'' is a precise term in the context of normalizing flows, but we stick with Gaussianization in accordance with the early literature.

Treating each dimension separately is motivated by an exact decomposition of the loss defined earlier in \cref{eq:loss}:
\eqall{
    \Ll &
    = \KL{q(z)}{\Nn(0, I)} \\&
    = \underbrace{\KL{q(z)}{q(z_1) \cdots q(z_D)}}_{\text{Dependence } \Dd} \\&\qquad
    + \sum_i \underbrace{\KL{q(z_i)}{\Nn(0, 1)}}_{\text{marginal loss } \Jj_i}.\!\!
}{\label{eq:pythagorean-theorem-dependence}}
This representation makes clear that the loss is composed of two parts: The \emph{dependence $\Dd$}, which measures how far $q(z)$ deviates from the product of its marginal $q(z_1) \cdots q(z_D)$, and the \emph{marginal loss} for each dimension, which measure the deviation of each marginal $q(z_i)$ from the univariate standard normal. Such a result is known as a generalized Pythagorean Theorem in information geometry \cite{amari_methods_2007}. \cref{eq:pythagorean-theorem-dependence} is due to \cite{cardoso_dependence_2003}.

A single layer can reduce the marginal losses $J_i$ in all dimensions (close) to zero if each $f_{\text{dim}, \theta_i}$ is sufficiently rich. This leaves only the dependence $\Dd$ as the loss. Note that fundamentally, the dependence $\Dd$ cannot be changed by a single-dimensional transformation.

If all layers worked with the same set of dimensions, the dependency $\Dd$ part of the loss would never change.
Here, the rotation $f_\text{rot}$ layer in each block comes into play (see \cref{eq:gaussianization-block}). It is parameterized by an orthogonal matrix $Q \in O(D)$ that rotates the data, varying the directions of action of the dimension-wise transforms:
\eql{
    f_\text{rot, Q}(x) = Qx.
}
Rotating the data does not affect the sum of the loss contributions $\Ll = \Dd + \sum_i \Jj_i$, but it redistributes it between the dependence $\Dd$ and the marginal losses $\Jj_i$. This makes choosing rotations $Q$ crucial. We elaborate on this issue in \cref{sec:rotations}.

Together, this model is sufficiently rich so that with random rotations and enough expressive layers, Gaussianization can approximate arbitrary distributions, i.e.~map any input dataset $p(x)$ to normally distributed codes \cite{chen_gaussianization_2000,meng_gaussianization_2020}. Note that the existing guarantees only consider weak convergence, whereas no convergence guarantee on the KL divergence $\Ll$ is known. In this work, we consider the KL divergence, since this stronger notion of convergence ensures that both samples and density estimates converge.

\subsection{Training}

Gaussianization can be trained layer-by-layer (\textit{iterative}) or by training all blocks jointly (\textit{end-to-end}).

In iterative training, one adds layers one by one: The data set is used to train the first block $f_\text{block}$ to maximally reduce the loss in \cref{eq:loss}. The second block is then trained with the data transformed by the first layer.

In end-to-end training \cite{meng_gaussianization_2020}, we would concatenate a pre-specified number of blocks at initialization. The parameters of each block are then trained jointly using \cref{eq:loss}. The advantage of end-to-end training is that blocks can collaborate as the training signal is the gradient of the entire pipeline and not just of a single block.

Iterative training has the advantage that there are fast approaches for learning the one-dimensional transforms. The cumulative distribution function (CDF) may be estimated on each 1D slice using the quantiles of the data, which by-passes gradient descent on the loss. For example, \citet{dai_sliced_2021} use rational-quadratic splines to fit the CDF due to their flexibility and analytical invertibility.

In practice, end-to-end training requires fewer layers than iterative training when learning a given distribution in low dimensions \cite{meng_gaussianization_2020}. As the dimension increases, however, the convergence of the training saturates. %
State-of-the-art Gaussianization results in high dimensions (MNIST, CIFAR10) are currently held by iterative training with SINF \cite{dai_sliced_2021}.

\subsection{Choosing rotations}
\label{sec:rotations}

Looking at the loss in \cref{eq:loss}, it becomes clear that the rotation layers have a strong impact on the performance. The KL divergence $\KL{q(z)}{\Nn(0, I)}$ itself is symmetric under rotations, and so the total loss $\Ll$ in \cref{eq:loss} does not change with the rotation. Instead, it distributes the loss between the dependence $\Dd$ and the marginal losses $\Jj_i$. To illustrate this, consider two special examples:

First, we take a distribution $p(x) \neq \Nn(0, I)$ which can be written as the product of marginals in some rotation of the data $Q^*$:
\eql{
    p(Q^*x) = p((Q^*x)_1) \dots p((Q^*x)_D).
}
If we evaluate the loss in this orientation, the dependence $\Dd$ becomes zero and all loss is contained in the marginals $\Jj_i$. This allows Gaussianization to fit $p(x)$ in one layer: The rotation layer chooses $Q = Q^*$ and the element-wise transforms fit $p((Q^*)x_i)$.

On the other end, consider a Gaussian distribution $p(x) = \Nn(0, \Sigma)$ where $\tr\Sigma = D$. Then, there exists a rotation $Q^+$ for which the standard deviations along the axis are one, i.e.~$(\Sigma')_{ii} = ((Q^+)^\transy \Sigma Q^+)_{ii} = 1$. Then, $\Jj_i = 0$ and all the loss is contained in the dependence $\Dd$. The element-wise transformation layer cannot make any progress in this situation.

For the two examples given, it is in principle known how to obtain the optimal choice for $Q$: In the case of the first distribution, ICA aims to find $Q^*$ such that the axes are independent. For the case of Gaussian distributions, PCA yields the optimal $Q^*$ such that $\Sigma'$ is diagonal and can be solved using one block.

For most real-world distributions, an orientation $Q^*$ does not exist where the data dimensions become independent.

The Cramér-Wold theorem \cite{cramer_theorems_1936} guarantees that the learned latent codes are exactly Gaussian if and only if there is no orientation $Q$ with a non-Gaussian marginal. To visualize this result, imagine projecting the data set along a unit vector and looking at the histogram:
\eql{
    z_\text{proj} = w^\transy z.
}
If the codes $z$ are distributed like a multivariate standard normal distribution, then each projection $z_\text{proj}$ will be distributed like a univariate standard normal distribution. If however, $z$ is not normally distributed, then there must be some projection for which the data is also not distributed like a normal distribution.

This implies that for a single layer we want to choose $Q$ such that the marginal projections are as non-Gaussian as possible. Then, as much loss as possible is contained in $J_i$, which can then be removed by the single-dimensional transformations $f_{\text{dim},\theta_i}$.

There is a rich history in identifying interesting marginal directions in high-dimensional data. These are the most common choices for computing $Q$ from data:
\begin{itemize}
    \item Random rotations are randomly sampled as $Q \in O(D)$. We give several guarantees in \cref{sec:convergence-guarantees} for this case.
    \item Principal Component Analysis (PCA) transforms any distribution with nontrivial covariance $\Sigma$ such that its principal axes coincide with the coordinate axes, i.e.~such that the resulting covariance matrix is diagonal $Q^\transy \Sigma Q = \Diag(S)$ with eigenvalues $S$.
    \item Independent Component Analysis (ICA) identifies the space in which a $p(u) = p(u_1) \cdots p(u_D)$, if $x = Au$ and $u$ can be factorized in this form.
    \item \maxKSWD{} identifies the directions in which the sliced Wasserstein distance can be maximally reduced. The sliced Wasserstein distances can be a proxy for the marginal KL divergences, as both measure a divergence between distributions.
\end{itemize}

We consider limitations of learned rotations in \cref{sec:guarantee-learned-rotations}.

\section{Analytic scaling behavior with dimension}
\label{sec:convergence-guarantees}

In this section, we derive the scaling behavior of Gaussianization with the dimensionality of $p(x)$: We show that the number of required layers grows at least linearly with the dimensionality of the problem in the case of random rotations. We adopt the notation for asymptotic convergence behavior with dimension $D$ from \cite{knuth1997art}, replacing $O \to \Oo$ to avoid name collision with the orthogonal group $O(D)$:
\eqal{
    \Oo(f(D)) &:= \{ g(D) \leq C f(D) \text{ for } D > D^* \}, \\
    \Omega(f(D)) &:= \{ g(D) \geq C f(D) \text{ for } D > D^* \}, \\
    \Theta(f(D)) &:= \Oo(f(D)) \cap \Omega(f(D)).
}

\subsection{Random rotations}
\label{sec:guarantee-random}

\begin{figure}
    \centering
    \includegraphics[width=.9\linewidth,trim={0 1.3cm 0 0},clip]{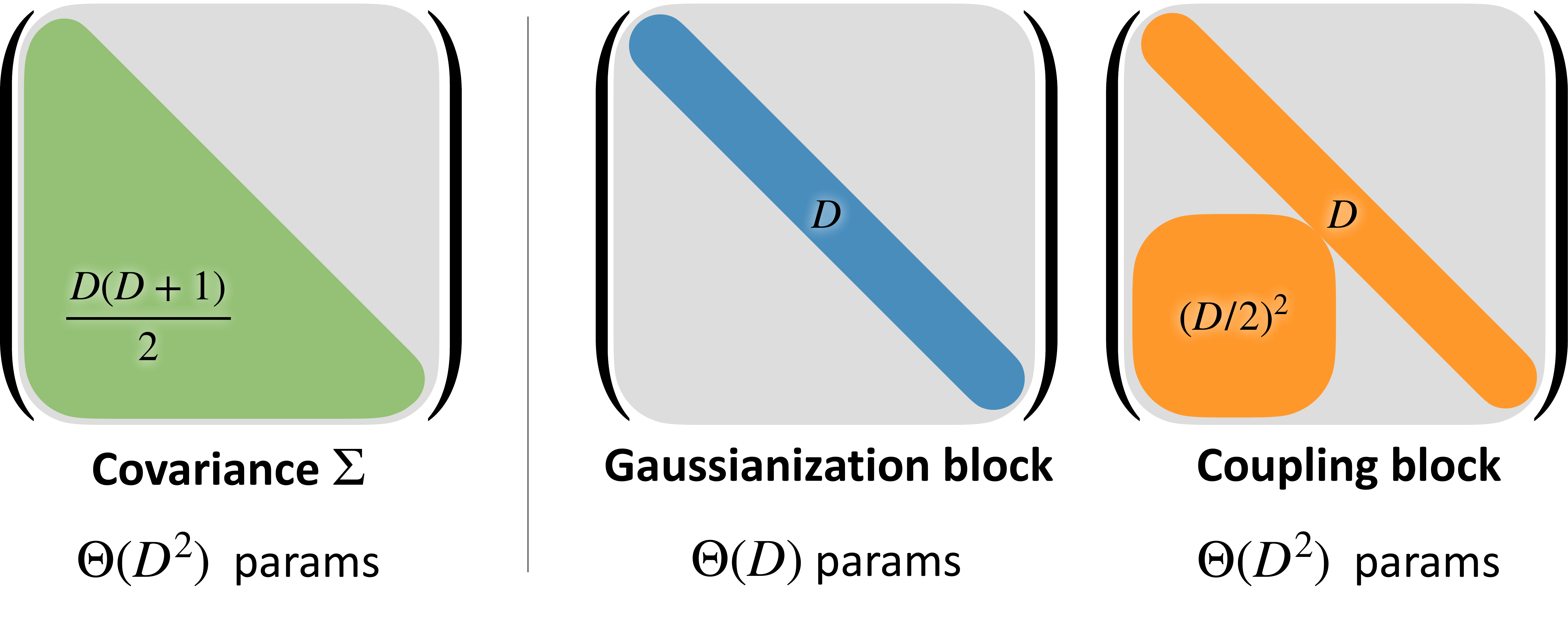}
    \caption{\textbf{Parameter counting argument}: The goal is to transform the covariance matrix $\Sigma$ to the unit matrix $I$. The covariance has $D(D+1)/2$ degrees of freedom, of which Gaussianization can learn $D$ per layer, and couplings $D^2/4 + D$ per layer.}
    \label{fig:gaussianization_effect}
\end{figure}

We compute the convergence scaling for the special case where the data distribution to be learned is a multivariate Gaussian $p(x) = \Nn(0, \Sigma)$. This may seem like a large restriction compared to the general distributions Gaussianization promises to learn. However, the goal of Gaussianization is to map the data to a standard normal, so after enough Gaussianization blocks the latent estimate will be close to a normal distribution $\Nn(0, \Sigma)$ with $\Sigma \approx I$. As we only give lower bounds on the number of required layers, our bounds will be correct even if coming close to this case is hard. %
Together, we expect that difficulties in learning simple distributions like Gaussians will also occur when learning complicated distributions. We provide experiments in \cref{sec:experiments} that show how the central insights transfer to arbitrary distributions.

\begin{assumption}
    \label{as:normalized-covariance}
    The covariance is normalized: $\tr\Sigma = D$.
\end{assumption}
\begin{assumption}
    \label{as:non-degenerate-eigenvalues}
    The eigenvalues of the covariance matrix $\Sigma$ are distinct: $\lambda_i \neq \lambda_j$ for $i \neq j$.
\end{assumption}
Both assumptions are weak: \cref{as:normalized-covariance} is a typical data preprocessing, and it is also achieved after a single Gaussianization layer. \cref{as:non-degenerate-eigenvalues} is typically satisfied when working with real data that are in `general position'. %

For Gaussian input, we find the following result:
\begin{theorem}
    \label{thm:end-to-end-scaling}
    Given a multivariate Gaussian distribution $p(x) = \Nn(0, \Sigma)$ under \cref{as:normalized-covariance,as:non-degenerate-eigenvalues}.
    To exactly represent $p(x)$ with \emph{random} rotations $Q$, at least 
    \eql{
        L \geq \frac{1}{2} (D + 1)
    }
    Gaussianization layers are required almost surely.
\end{theorem}
This is a lower bound on the required number of layers for both the end-to-end and iterative training approaches: \textbf{The number of Gaussianization blocks required grows (at least) linearly with the number of dimensions.} The proof is a simple parameter-counting argument: $\Sigma$ has $D(D+1)/2$ degrees of freedom, but linear Gaussianization with random rotations only has $D$ parameters per layer. Dividing $D(D+1)/2$ by $D$ yields the result. This is visualized in \cref{fig:gaussianization_effect}. In the proof in \cref{app:proof-end-to-end-scaling} we give more details on the role of rotations.

The above argument assumes that we want exactly represent the target distribution $p(x) = \Nn(0, \Sigma)$. This is an unrealistic requirement for machine learning models: Imagine training a neural network with ReLU activations to fit a parabola $f(x) = x^2$. Being piecewise linear, you can never hope for the ReLU network to exactly represent the parabola. Instead, one requires that some error metric between the learned and the true function can be made arbitrarily small. The existence of arbitrarily close models is called \textit{universal approximation}.

In this work, we consider the loss that is used in practice: the Kullback-Leibler divergence in \cref{eq:loss}, which is equivalent to the maximum likelihood criterion. It is a strong metric in the sense that it ensures convergence of both samples (``$x \sim q \to x \sim p$'') and the corresponding densities (``$q \to p$''). For the Gaussian case, the KL divergence between some $p(x) = \Nn(0, \Sigma)$ and the latent $\Nn(0, I)$ reads:
\eqall{
    \Ll^{(0)} &
    = \KL{\Nn(0, \Sigma)}{\Nn(0, I)} \\&
    = \tfrac12 (\tr \Sigma - D - \log \det \Sigma) \\&
    \overset{\!\!\text{(\tiny A\ref{as:normalized-covariance})}\!\!}{=} -\tfrac12 \log\det\Sigma.
}{\label{eq:loss-gaussian}}
This holds under the weak \cref{as:normalized-covariance} that $\tr\Sigma = D$.

Universal approximation results are often pure existence theorems: They only state that a suitable network can be built, but often do not tell anything about the required model complexity. The existing universal approximation guarantees for Gaussianization \cite{chen_gaussianization_2000,meng_gaussianization_2020} also do not provide an explicit guarantee for the required model capacity. In addition, they do not consider the KL divergence, but weak convergence, which does not imply convergence of densities.

For the first time, we give a rigorous lower bound on the KL divergence assuming a given number of layers for random rotations $Q \in O(D)$.
\begin{theorem}
    \label{thm:iterative-convergence-rate}
    Given a multivariate Gaussian distribution $p(x) = \Nn(0, \Sigma)$ under \cref{as:normalized-covariance}. Then, in the case $\Ll \ll 1$, the expected loss after $L$ iterative Gaussianization blocks with random rotations is approximately:
    \eql{
        \label{eq:iterative-convergence-rate}
        \EE_{Q_{1 \dots L} \in O(D)}[\Ll^{(L)}] \leq \left(1 - \frac{2}{D + 2}\right)^L \Ll.
    }
\end{theorem}
\cref{thm:iterative-convergence-rate} sharpens \cref{thm:end-to-end-scaling} to a statement on the KL divergence. Rewriting \cref{eq:iterative-convergence-rate} to derive how many layers $L$ are required at least to reduce some initial loss $\Ll$ to $\Ll'$, we find the same scaling behavior with the dimension $D$ as in \cref{thm:end-to-end-scaling}:
\eql{
    \label{eq:iterative-scaling}
    L \geq \frac{\log(\Ll'/\Ll)}{\log\!\big(1 - \frac{2}{D+2}\!\big)} \approx \log(\Ll/\Ll') \frac{D+1}{2} = \Omega(D).
}
In words: The number of layers $L$ required to reduce the loss by a fixed amount scales at least linearly with the dimension $D$. For the proof, see \cref{app:iterative-convergence-rate}.

\cref{fig:gaussian-scaling-experiment} empirically confirms this theoretical bound on a variety of normal distributions with different initial $\Sigma$. It confirms the scaling with $\Omega(D)$ and shows that the actually required numbers to reduce the loss by a ratio is higher than \cref{eq:iterative-scaling}.
In order to be consistent with $\Ll \ll 1$, we measure the number of required layers for each input by extrapolating the convergence rate of the last two of $10D$ layers. If $\Ll \gg 0$, the performance deviates from the exact prediction in \cref{eq:iterative-scaling} in both directions, but the scaling with dimension is preserved. Details are given in \cref{app:gaussian-experiment}.

While \cref{thm:iterative-convergence-rate} only holds for the iterative case, end-to-end training may perform better. We address this in \cref{cor:end-to-end-scaling-learned-rotations} in the following section.

To the best of our knowledge, \cref{thm:end-to-end-scaling,thm:iterative-convergence-rate} are the first explicit convergence results for Gaussianization.

\subsection{Limitations of learned rotations}
\label{sec:guarantee-learned-rotations}

In the previous section, we showed the scaling behavior of Gaussianization for random rotations. In this section, we show that end-to-end training has the same scaling behavior with dimension. For iterative training, we point at a fundamental challenge in learning high-dimensional rotations from data.

End-to-end training with \textit{learned} rotations may outperform \cref{thm:end-to-end-scaling} in terms of the number of required layers. In fact, rotations exist such that arbitrary $\Sigma$ can be fit with a single layer. However, state-of-the-art rotation learning typically does not train on the full orthogonal group $O(D)$, as current methods become prohibitively expensive with increasing $D$. Instead, one considers subsets typically spanned by $k \cdot D$ independent parameters, such as spanned by Householder transforms $I - vv^\transy$ ($k=1$) or block-diagonal orthogonal matrices (with $k=(b-1)/2$ for block size $b$) \cite{meng_gaussianization_2020}. Even with these parameterized rotations, however, the number of required layers scales with the dimension $D$. We then find:
\begin{corollary}
    \label{cor:end-to-end-scaling-learned-rotations}
    Given a multivariate non-degenerate Gaussian distribution $p(x) = \Nn(0, \Sigma)$ under \cref{as:normalized-covariance,as:non-degenerate-eigenvalues}.
    To exactly represent $p(x)$ with \emph{learned} rotations $Q$ with $k \cdot D$ parameters each, at least
    \eql{
        L \geq \frac{1}{2(k+1)} D
    }
    Gaussianization layers are required almost surely.
\end{corollary}
The scaling of $L$ with $D$ can only be avoided when $k \in \Omega(D)$, which does not hold for the parameterizations mentioned above.

For iterative training, the bulk of the literature focuses on rotating the data such that the marginals deviate as much as possible from the target normal distribution $\Nn(0, 1)$, see \cref{sec:rotations}. 
Alternative methods differ mainly by their measure of marginal non-Gaussianity.

Unfortunately, this approach has an inherent tendency to overfit on finite training sets: 
In the non-asymptotic regime, there is a high probability that some marginal projections exhibit considerable spurious non-Gaussianity even when the data are sampled from a perfect standard normal, $x \sim \Nn(0, I)$.
In other words, although training has converged, the iterative training algorithm will identify a rotation $Q$ that appears to improve the loss but actually worsens it.
In sufficiently high dimensions $D$, this still happens for large datasets with $N \gg D$, a typical situation in computer vision.
Inspired by \citep{bickel_projection_2018}, we illustrate the phenomenon at $D=3072=32 \times 32 \times 3$ and $N=60000$, the dimension and size of the CIFAR10 dataset.

\begin{figure}
    \centering
    \includegraphics[width=.8\linewidth]{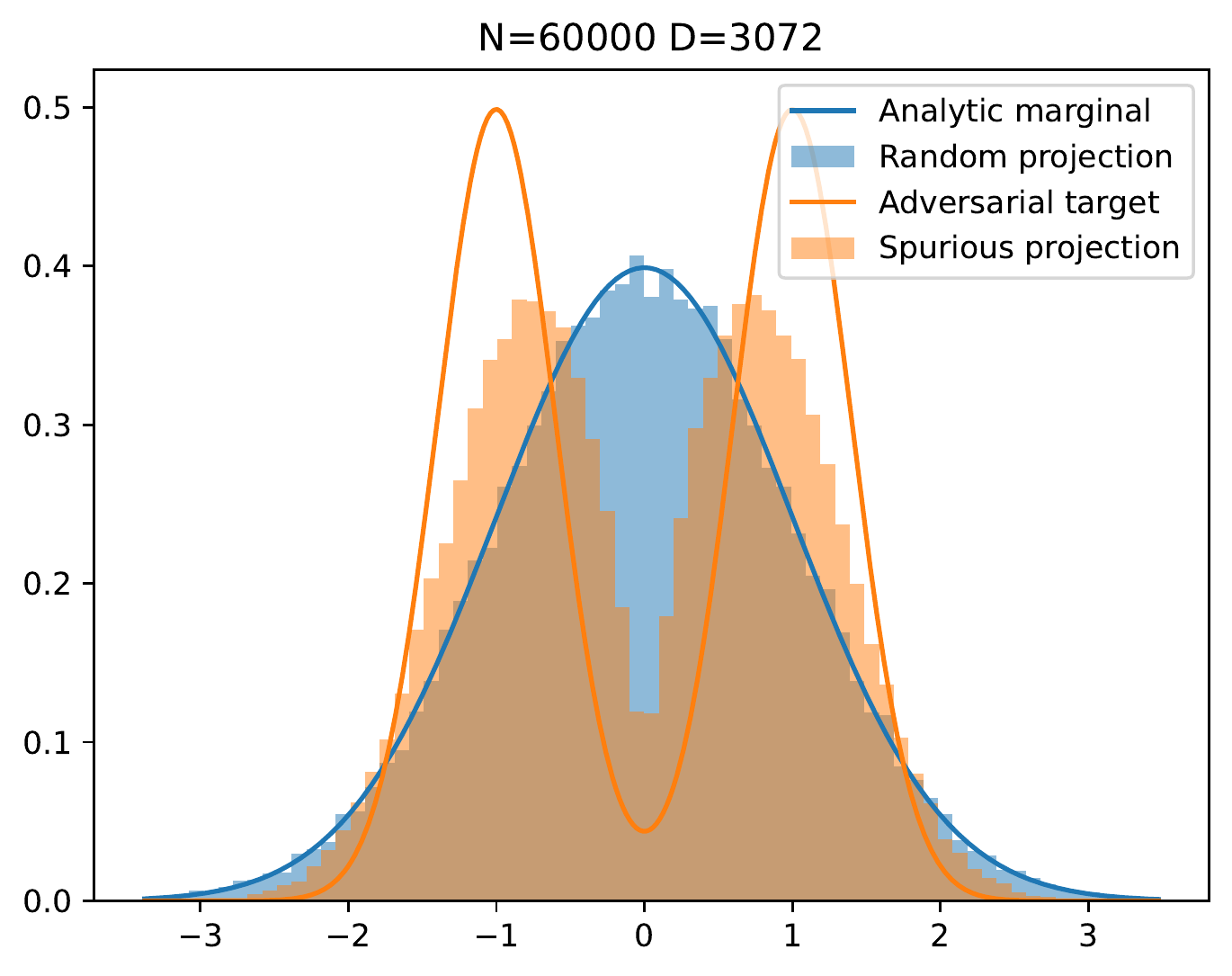}
    \caption{\textbf{Spurious projection of standard normal data}. The plot shows $N=60,000$ samples from a $D=3072$-dimensional standard normal distribution projected to a single dimension. The {\color{C0}blue} projection is selected randomly, and the resulting histogram is close to standard normal. The {\color{C1}orange} projection is optimized so that the dataset has a spurious bimodal histogram. The histograms coincide with the marginal distribution Gaussianization would learn, producing a bimodal distribution from Gaussian data in the second case.}
    \label{fig:spuriously-non-gaussian}
\end{figure}

In \cref{fig:spuriously-non-gaussian}, we show the histogram of a spurious non-Gaussian projection
\eql{
    x_\text{proj} = w^\transy x, \qquad |w|=1, x \sim \Nn(0, I)
}
of $N$ fixed samples from a $D$-dimensional standard normal.
We construct $w$ such that the projection of the fixed data is as close as possible to the bimodal distribution shown in the plot. 
Although in the asymptotic limit $N \to \infty$ no such $w$ exists, the optimization readily finds a solution in the finite dataset.
Details on the experiment can be found in \cref{app:spurious-rotations}.

A similar experiment is reported in Appendix D.1 of \cite{dai_sliced_2021}. They show that \maxKSWD{} can identify spurious non-Gaussian projections. We extend their experiment by demonstrating the effect of such projections on the learned distribution. In our example, Gaussianization would fit a bimodal distribution to a standard normal.

The theoretical analysis in chapter 8 of \cite{wainwright_high-dimensional_2019} shows that phenomena like this are fundamental in finite datasets.
Specifically, they prove that \textit{no} method can reliably estimate the eigenvectors of the \textit{empirical} covariance in high dimensions  if the ratio $D/N$ is bounded away from zero, unless additional assumptions (e.g. sparsity of the covariance) are made.
Consequently, even the straightforward idea of defining the optimal $Q$ via PCA can fail and must be used with caution.

In practice, we expect Gaussianization with learned rotations to work well in the initial blocks, where the intermediate latent distributions are strongly non-Gaussian.
Deeper in the network, however, intermediate distributions are already close to standard normal, and spurious projections will appear.
The resulting overfit of the rotation $Q$ will fool the subsequent marginal transformation, making the data less rather than more Gaussian.
This can only be fully avoided by random orthogonal matrices $Q$, which almost surely do not result in spurious non-Gaussian projections \cite{bickel_projection_2018}, compare the random projection in \cref{fig:spuriously-non-gaussian}. 
Random rotations are exactly the regime of our theoretical results.

\subsection{Relation to coupling-based normalizing flows}
\label{sec:coupling-comparison}

We now compare the above scaling results to guarantees for another normalizing flows architecture, namely all variants of coupling-based normalizing flows \cite{dinh_nice_2015}. See \citet{draxler_whitening_2022} for a list of architectures the following results apply to.

Coupling-based normalizing flows and Gaussianization (see \cref{sec:setup}) both represent invertible functions as a composition of invertible blocks, and each block consists of a rotation followed by a reshaping of the distribution. 
Each coupling-based reshaping splits the data dimensions into a $D'$-dimensional active and $(D-D')$-dimensional passive subspace, and reshapes the active subspace {\em conditional} on the passive subspace.
Thus, couplings explicitly reduce the {\em dependency} between the active and passive dimensions, in contrast to Gaussianisation blocks, which reshape each dimension independently.
The downside of this improvement is that only $D'$ dimensions are modified in each block, with $D'=D/2$ being the most common choice.

Using exactly the same argument as in \cref{thm:end-to-end-scaling}, we provide a first glance on how many coupling blocks are required to fit $p(x) = \Nn(0, \Sigma)$:
\begin{corollary}
    \label{thm:end-to-end-scaling-coupling}
    Given a multivariate Gaussian distribution $p(x) = \Nn(0, \Sigma)$.
    To exactly represent $p(x)$, at least
    \eql{
        L_{\mathrm{cpl}} \geq 2 \in \Omega(1)
    }
    coupling blocks with random rotations $Q \in O(D)$ are required almost surely.
\end{corollary}
This result says that the number of coupling blocks required to represent $\Nn(0, \Sigma)$ exactly is independent of dimension in the best case. For the proof, see \cref{app:end-to-end-scaling-coupling}.

Just like \cref{thm:end-to-end-scaling-coupling} gives a lower bound on the number of blocks required, Theorem 2 of \cite{koehler_representational_2021} provides an upper bound, which we condense for simplicity:
\begin{theorem}[\citet{koehler_representational_2021}]
    \label{thm:end-to-end-scaling-coupling-48}
    Given a multivariate Gaussian distribution $p(x) = \Nn(0, \Sigma)$.
    To exactly represent $p(x)$, at most $L_{\mathrm{cpl}} \leq 48 \in \Oo(1)$ coupling blocks with block permutations are required.
\end{theorem}
This ensures that at most 48 blocks or $\Oo(1)$ are required to exactly fit a Gaussian $p(x) = \Nn(0, \Sigma)$. While this may seem like a large number, the result crucially ensures that the number of required blocks is independent of the dimension. In the statement, block permutations refers to all rotations switching all active and passive dimensions:
\eql{
    Q = \begin{bmatrix}
        0 & I_{D/2} \\
        I_{D/2} & 0
    \end{bmatrix}.
}
Odd and even layers each modify one half of the dimensions respectively. This is a popular choice in practice.

The above arguments considered exact representation of $p(x)$, and statements about the convergence rate are stronger as argued in \cref{sec:guarantee-random}. We therefore condense (see \cref{app:iterative-convergence-rate-coupling}) the main theorem in \cite{draxler_whitening_2022} to have the same format as \cref{thm:iterative-convergence-rate}:
\begin{theorem}[\citet{draxler_whitening_2022}]
    \label{thm:iterative-convergence-rate-coupling}
    Given a multivariate Gaussian distribution $p(x) = \Nn(0, \Sigma)$. The initial loss $\Ll$ is given by \cref{eq:loss-gaussian}.
    Then, in the case $\Ll \ll 1, 1 \ll D$, the loss after $L_{\mathrm{cpl}}$ iterative coupling blocks with random rotations is at most:
    \eql{
        \label{eq:iterative-convergence-rate-coupling}
        \EE_{Q_{1 \dots L} \in O(D)}[\Ll^{(L_{\mathrm{cpl}})}] \lesssim \left(\frac12 \right)^{L_{\mathrm{cpl}}} \Ll.
    }
\end{theorem}
In parallel to the previous calculations, we derive how many coupling blocks are required to reduce the loss from $\Ll$~to~$\Ll'$:
\eql{
    \label{eq:iterative-coupling-scaling}
    L_{\mathrm{cpl}} \lesssim \frac{\log(\Ll/\Ll')}{\log(2)} = \Oo(1).
}

Together, \cref{thm:end-to-end-scaling-coupling} from below and \cref{thm:end-to-end-scaling-coupling-48,thm:iterative-convergence-rate-coupling} from above show that in contrast to Gaussianization, the number of coupling blocks is independent of the problem dimension:
\eql{
    L_{\mathrm{cpl}} = \Theta(1).
}
This means that on Gaussian data, \textbf{Gaussianization requires more layers than coupling blocks because the layers do not model dependencies between dimensions.}

\newpage

\section{Experimental scaling behavior}
\label{sec:experiments}

In \cref{sec:convergence-guarantees}, we gave rigorous predictions for the convergence of Gaussianization on Gaussian input $p(x) = \Nn(0, \Sigma)$. The core result is that the number of blocks increases linearly with the dimension. Comparing this result to similar results on coupling-based normalizing flows, we identified that modeling dependencies between dimensions is the major bottleneck of Gaussianization.

We now lift the restriction to Gaussian input and consider $p(x) \neq \Nn(0, \Sigma)$. We find that as the dimension increases, the number of required layers $L$ increases with dimension $D$, but favorable scaling can be achieved depending on the properties of the data.
We give implementation details in \cref{app:gaussianization-implementation}. We base our implementation of $f_{\operatorname{dim}}(x)$ on the code provided by SINF \cite{dai_sliced_2021}.

\subsection{Toy scaling experiment}
\label{sec:dependency-toy}

\begin{figure}[bt]
    \centering
    \includegraphics[width=.9\linewidth]{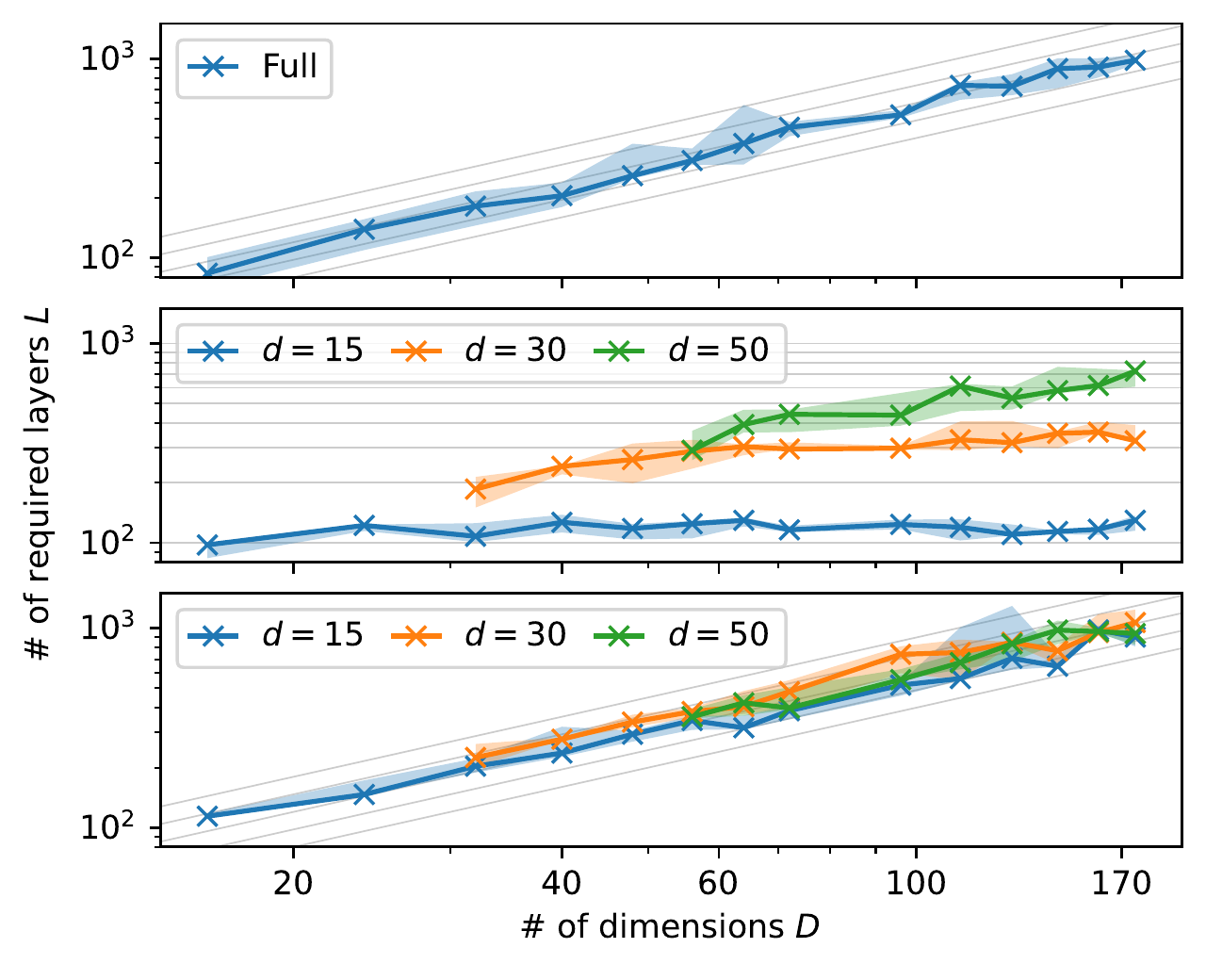}
    \caption{\textbf{Required layers of Gaussianization on toy data.} \textit{(Top)}~If all dimensions depend on one another, the number of required layers increases linearly with dimension. \textit{(Middle)}~If trailing dimensions $i > d$ are pairwise independent given the core $d$ dimensions, only about a constant number of layers is sufficient for fixed $d$. \textit{(Bottom)}~If the trailing dimensions $i > d$ are independent Gaussian noise, the number of layers increases linearly with dimension. Shaded regions indicate 100\% of the training runs. Gray lines indicate $\Theta(D)$ resp.~$\Theta(1)$.}
    \label{fig:scaling-laws}
\end{figure}

For determining the scaling behavior of Gaussianization we consider a family of distributions of varying dimension $D$. We propose to build such a toy distribution autoregressively:
\eql{
    \label{eq:toy-distribution-form}
    p(x) = p(x_1) \prod_{i=2}^D p(x_i | A_i),
}
where the set $A_i \subseteq \{ x_1, \dots, x_{i-1} \}$ collects the random variables that $x_i$ depends upon. This allows adding new dimensions by specifying their dependencies.

We consider the following three variants: (1) Let every variable depend on all previous variables: $A_{i}^{(1)} = \{ x_1, \dots, x_{i-1} \}$). (2) We only make a subset of $d$ variables depend on all previous, and let the remaining dimensions depend on this fixed subset of dimensions: $A_{i \leq d}^{(2)} = \{ x_1, \dots, x_{i-1} \}$ (core) and $A_{i > d}^{(2)} = \{ x_1, \dots, x_d \}$ (remainder). (3) Like the second case, but the remaining dimensions $i>d$ are independent Gaussian noise: $A_{i \leq d}^{(2)} = A_{i \leq d}^{(3)}$ (core) and $A_{i > d}^{(3)} = \emptyset$ (noise).

In particular, we choose $p(x_i|A_i)$ as a continuous mixture of Gaussians
\eqall{
    p(x_1) &= \Nn(m_1, \sigma_1^2), \\
    p(x_i|A_i) &= \Nn(m_i(A_i), \sigma_2^2)
}{\label{eq:toy-distribution-details}}
where the dependencies are introduced through $m_i(A_i)$:
\eql{
    \label{eq:toy-distribution-continuous-mixture}
    m_i(A_i) = m_0 + 5 \tanh\!\bigg(\frac{1}{10} \sum_{x_j \in A_i} s_{ij} x_j^2\bigg).
}
The values $m_1, m_0 \in \RR; \sigma_1, \sigma_2 \in \RR_+, s_{ij} \in \{-1, 1\}$ are parameters to the distribution. %

\cref{fig:scaling-laws} shows how many layers~$L$ are needed to reduce the loss by a fixed ratio $\gamma = \Ll' / \Ll < 1$ for each case as a function of dimension~$D$.
We find that for cases (1) and (3), the number of required Gaussianization layers increases linearly with dimension, which is consistent with our theoretical result on Gaussian data in \cref{eq:iterative-scaling}. In case (2) however, the number of required layers remains roughly constant with dimension (but it does depend on the number of dependent dimensions).\footnote{In a preliminary version of this paper, we conjectured that \textit{alone the number} of dependencies causes the scaling behavior with dimension. After more detailed experiments, this hypothesis turned out to be false.} We show in \cref{fig:toy-random-projections} in \cref{app:dependency-toy} that random projections are less Gaussian in this case, as additional variables carry information about the core dimensions. This makes it easier for Gaussianization to fit the data, efficiently removing loss by fitting the non-Gaussian marginals.

This toy experiment indicates that linear increase in required layers holds for some distributions, and a favorable scaling behavior may be obtained for certain input.

\subsection{Real dataset experiment}
\label{sec:dependency-real}

\begin{figure}[tb]
    \centering
    \includegraphics[width=.9\linewidth]{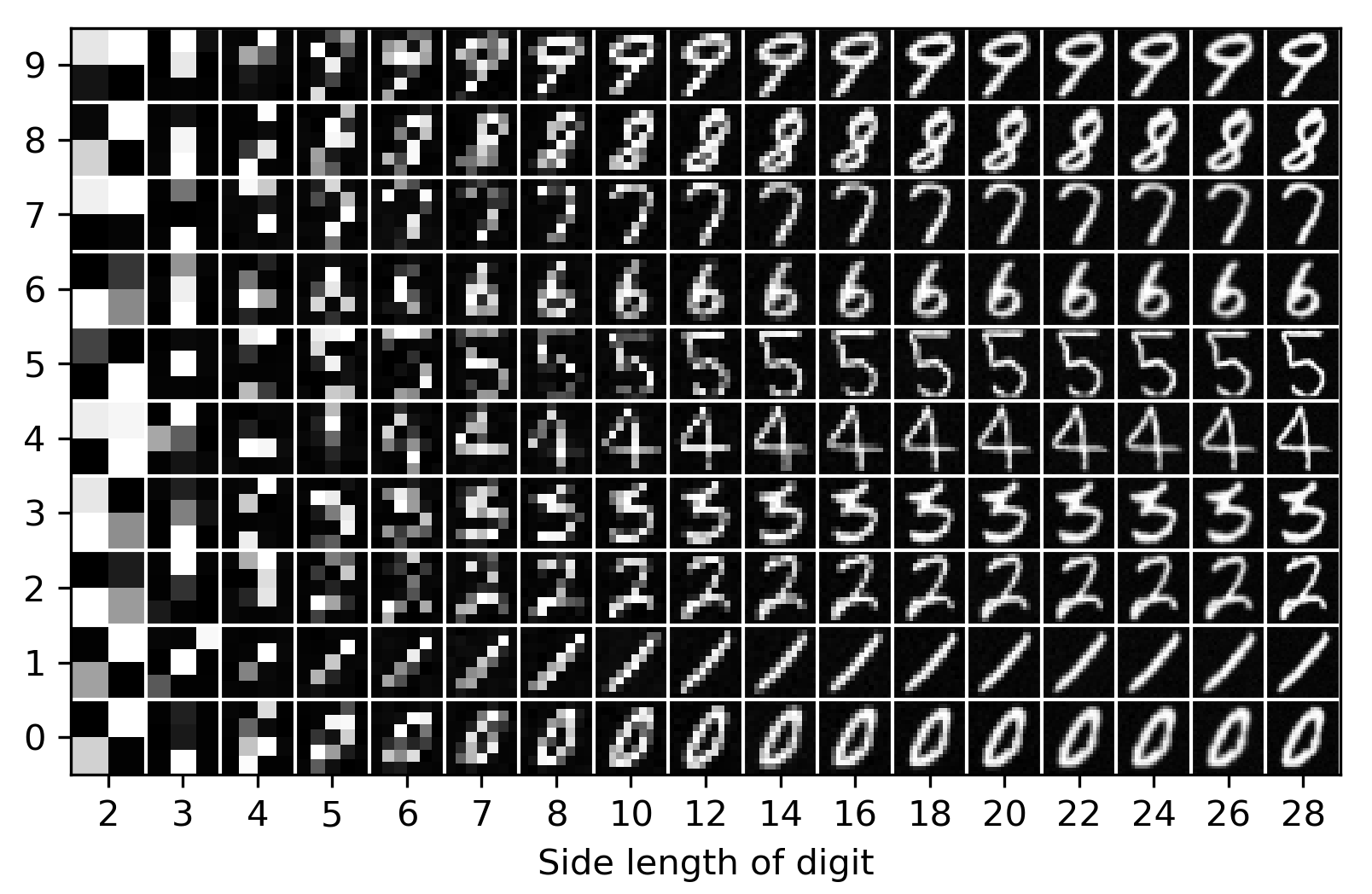}
    \caption{Our multi-scale EMNIST digits dataset.}
    \label{fig:emnist-rescaled}
\end{figure}

We now consider the scaling behavior of Gaussianization on a real dataset, the EMNIST digits \cite{cohen2017emnist}.
To measure the scaling with dimensions, we construct variants of the data with different dimensions. We therefore rescale the images to scales between $2 \times 2$ and the original $28 \times 28$, see \cref{fig:emnist-rescaled}.

\begin{figure}[tb]
    \centering
    \includegraphics[width=.9\linewidth]{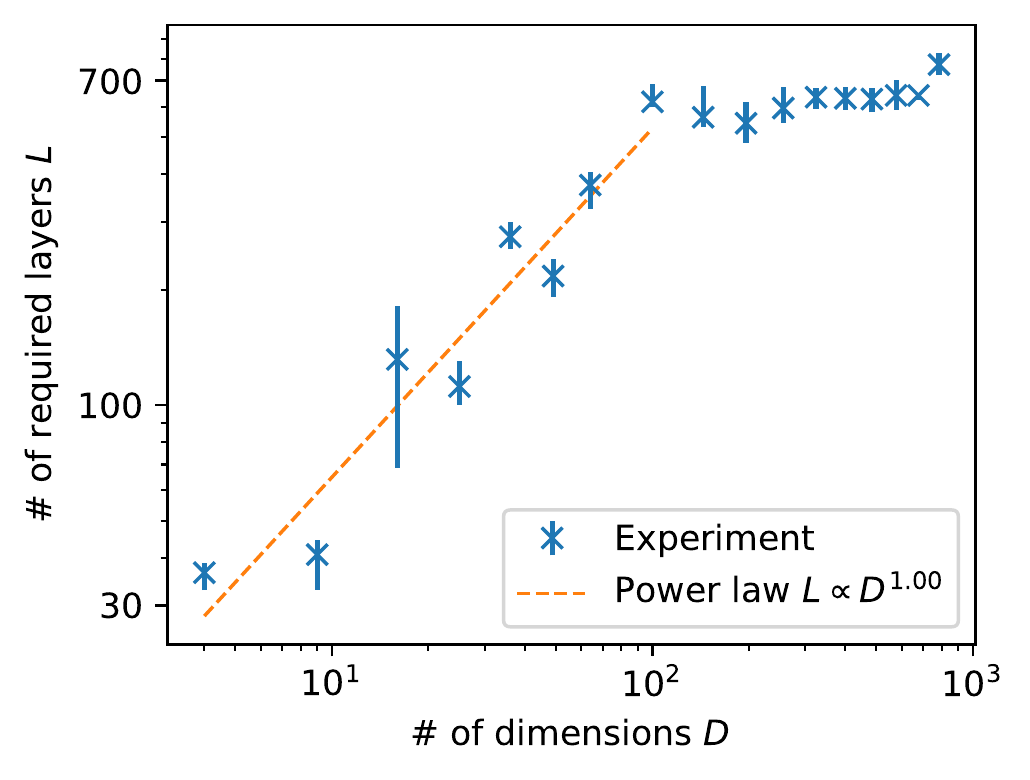}
    \caption{\textbf{Gaussianization requires more layers for higher resolution datasets.} From $2 \times 2$ to $10 \times 10$, fitting a power law yields a linear function. For larger images, the difficulty does not increase further. We think that this is due to an increasing number pixels being largely determined from the other dimensions. Data points show the median and error bars cover 90\% of the training runs.}
    \label{fig:emnist-convergence}
\end{figure}

\cref{fig:emnist-convergence} shows how many layers are required as a function of dimension, extrapolated from training 64 Gaussianization layers. We find that up to a scale of $D=10 \times 10$, the number of required layers roughly increases as $\Theta(D)$, like in \cref{eq:iterative-scaling}, and then remains about constant.

Consistently, around a side length $l = 10$, the main characteristics of each digit become clear. Afterwards, only local details are filled in. We identify the corresponding scaling of Gaussianization with case (2) in \cref{sec:dependency-toy}, where additional dimensions are highly correlated with others, so that random marginal distributions become less Gaussian.

Note that for computing the absolute KL divergence as defined in \cref{eq:loss}, we need to evaluate the entropy of the data $H[p(x)]$. This is an unknown value in general and for rescaled EMNIST digits in particular. We therefore replace the ground truth dataset by generative models trained on each respective scale.
We use a Normalizing Flow for our ground truth distribution, which achieves better density estimates in general than Gaussianization. This removes bias from our convergence rate estimates: If we were to use the trained model itself as ground truth, we would find spuriously fast convergence even if the model does not converge -- as we compute the convergence rate to a suboptimal optimum. See \cref{app:dependency-real} for all experimental details.

\section{Conclusion}
\label{sec:Conclusion}

Gaussianization is a simple generative model, that has shown advantages in situations with little data \cite{meng_gaussianization_2020,dai_sliced_2021}. Differently from other approaches, it can be trained without loss and neural networks, yet provides useful density estimates and samples in low and moderate dimensions.

Scaling Gaussianization to high dimensions remains a major challenge, and we confirm this rigorously. We show analytically for the Gaussian distribution $p(x) = \Nn(0, \Sigma)$ (see \cref{sec:convergence-guarantees}) that the number of required layers typically scales with the dimensionality of the problem if the rotations $Q$ are chosen at random. On non-Gaussian distributions, we show that convergence can be favorable for some distributions. 

Generally speaking, our work complements the theory on networks that are constrained in order to be invertible: Should we have few unconstrained yet expensive layers (i.e.~autoregressive, with all dependencies modeled in a single, highly expressive layer \cite{knothe_contributions_1957,
rosenblatt_remarks_1952}) -- or should we constrain each layer, but have more of them (i.e.~coupling or Gaussianization layers)? Our theory indicates that Gaussianization, limited to learning marginals in each step, scales slower than alternatives, in particular in the regime of low loss.

An important point remains open, however: Improving Gaussianization to high dimensions may be possible by constructing better rotations $Q$. Our work points out fundamental limits: There exist spurious non-Gaussian directions that may be spuriously identified (see \cref{sec:guarantee-learned-rotations}), and we can never expect that a single Gaussianization block can fit real-world distributions (see \cref{sec:rotations}).

\section*{Acknowledgements}

\affSTRUCTURES{work}
AR acknowledges funding from the Carl-Zeiss-Stiftung. 
JM was supported by Informatics for Life funded by the Klaus Tschira Foundation. 
We thank Chenlin Meng, Biwei Dai, and Peter Sorrenson for the fruitful discussions, and the reviewers for their helpful comments.

\bibliography{bibliography}
\bibliographystyle{icml2023}

\newpage
\appendix
\onecolumn

\section{Proofs}
\subsection{Proof of \cref{thm:end-to-end-scaling}}
\label{app:proof-end-to-end-scaling}

\begin{proof}
    We need to exclude the special case that the accumulated rotation $Q^{(l)} = Q_l \cdots Q_1$ (partially) aligns with the eigenspace of $\Sigma$ for some $l = 1, \dots, L$. Then, the $\Sigma$ is (partially) diagonal in the input to this block $l$ and $f_\text{dim}$ is able to map $\Sigma \to I$ already earlier. However, this alignment has probability mass zero under random $Q$.
    
    To perfectly map the input Gaussian $\Nn(0, \Sigma)$ to the latent distribution $\Nn(0, I)$, we need to learn a linear function $A$ such that $A^\transy \Sigma A = I$. As a lower bound on how many layers we need to represent a suitable $A$, our learned function needs to have at least as many degrees of freedom as the covariance matrix $\Sigma  \in \RR^{D \times D}$: As it is symmetric $\Sigma$ has $D(D+1)/2$ independent degrees of freedom. The $D$ linear single-dimensional transforms in each block $f_{\text{dim},i}$ have a total of $D$ degrees of freedom. Thus we need more than $D(D+1)/(2D) = (D+1)/2 \geq D/2$ layers to represent $\Sigma$.
\end{proof}

\subsection{Proof of \cref{cor:end-to-end-scaling-learned-rotations}}
\label{app:proof-end-to-end-scaling-learned-rotations}

\begin{proof}
    The proof follows by replacing the number of parameters per layer in \cref{app:proof-end-to-end-scaling} by $(k+1)D$ instead of $D$. We find at least $D(D+1)/(2(k+1)D) = (D+1)/(2(k+1)) \geq D / (2(k+1))$ layers to represent $\Sigma$.
\end{proof}

\subsection{Proof of \cref{thm:iterative-convergence-rate}}
\label{app:iterative-convergence-rate}

\begin{proof}
    We start with a covariance matrix $\Sigma$ with $\tr\Sigma = D$ by \cref{as:normalized-covariance}.
    Given a fixed rotation $Q$, the output of the rotation layer is:
    \eql{
        p(Qx) = \Nn(0, Q \Sigma Q^\transy).
    }
    The marginals of this distribution read:
    \eql{
        p((Qx)_i) = \Nn(0, (Q \Sigma Q^\transy)_{ii}).
    }

    Each marginal can be transported immediately to a univariate standard normal via:
    \eql{
        f_{\text{dim}, \theta_i}(x_i) = \frac{x_i}{\sqrt{(Q \Sigma Q^\transy)_{ii}}}.
    }

    This makes the collected action of the element-wise layer:
    \eql{
        f_{\text{dim}}(x) = \Diag(Q \Sigma Q^\transy)^{-1/2} x =: S^{-1/2} x.
    }
    Here, $S = \Diag(Q \Sigma Q^\transy)$ collects the diagonal of the rotated covariance.

    The output of the layer is again a Gaussian distribution:
    \eql{
        \label{eq:gaussianization-block-on-gauss}
        p(x') = \Nn(0, \Sigma'), \quad \Sigma' = S^{-1/2} Q \Sigma Q^\transy S^{-1/2}.
    }
    Inserting this into the loss in \cref{eq:loss-gaussian}, we find:
    \eqal{
        \Ll'  &
        = -\tfrac12 \log \det \Sigma' \\&
        = -\tfrac12 \log \det (S^{-1/2} Q \Sigma Q^\transy S^{-1/2}) \\&
        = -\tfrac12 (\log \det S + \log \det \Sigma) \\&
        = \Ll -\tfrac12 \log \det S.
    }

    As we consider random rotations $Q \in O(D)$, we compute the expected loss over rotations:
    \eqal{
        \EE_{Q \in O(D)}[\Ll'] &
        = \Ll - \tfrac12 \EE_{Q \in O(D)}[\log \det S] \\&
        = \Ll - \tfrac12 \sum_{i=1}^D \EE_{Q \in O(D)}[\log (Q \Sigma Q^\transy)_{ii}] \\&
        \leq \Ll.
    }
    Here, we have used Jensen's inequality and that $\EE_{Q \in O(D)}[(Q \Sigma Q^\transy)_{ii}] = \tr\Sigma / D = 1$.
    (This is a vacuous bound: The expected loss after the layer is at least as good as before the layer.)

    We can estimate the error of Jensen's inequality to get an estimate for a lower bound on $\EE_{Q \in O(D)}[\Ll']$. Here, we make use of the result in \cite{costarelli_how_2015} applied to $\log$:
    \eql{
        \EE[\phi(x)] - \phi(\EE[x]) \leq \frac{1}{2} \max_{x \in I} \phi''(x) \Var[x].
    }
    In our case, $\phi(x) = -\log x$, and $x \in [\lambda_{\min}, \lambda_{\max}]$, the extremal eigenvalues of $\Sigma$ (which are invariant under rotation).

    In \cite{draxler_whitening_2022}, the authors show that:
    \eql{
        \Var[(Q \Sigma Q^\transy)_{ii}] = \frac{2}{(D + 2)} \Var[\lambda],
    }
    for the variance of the eigenvalues of $\Sigma$, given by $\Var_i[\lambda_i]$.

    For us, this means:
    \eql{
        \Ll - \Ll' \leq \frac{1}{2 \lambda_{\min}^2} \frac{2}{(D + 2)} \Var[\lambda]. 
    }

    We make use of following arithmetic mean-geometric mean (AM-GM) inequality by \cite{cartwright_refinement_1978}:
    \eql{
        \label{eq:am-gm-inequality}
        \frac{\Var[\lambda]}{2 \lambda_{\max}} \leq \bar\lambda - g \leq \frac{\Var[\lambda]}{2 \lambda_{\min}},
    }
    where $g$ is the geometric mean of the eigenvalues:
    \eql{
        g := \left( \prod_{i=1}^D \lambda_i \right)^{1/D},
    }
    and find:
    \eql{
        \Ll - \Ll' \leq \frac{1}{2 \lambda_{\min}^2} \frac{2}{(D + 2)} 2 \lambda_{\max} (1 - g).
    }

    As $\lambda_{\max} > 1$, we can multiply the right hand side by $\lambda_{\max}$:
    \eql{
        \Ll - \Ll' \leq \frac{1}{2 \lambda_{\min}^2} \frac{2}{(D + 2)} 2 \lambda_{\max}^2 (1 - g).
    }
    Then, rewrite using the conditioning number $\kappa = \lambda_{\max} / \lambda_{\min}$:
    \eql{
        \Ll - \Ll' \leq \frac{2}{(D + 2)} \kappa^2 (1 - g).
    }

    Note that one can write the loss $\Ll$ directly via $g$ and vice versa:
    \eqal{
        \Ll &= -\frac12 \log g^D = -\frac{D}{2} \log g, \\
        g &= \exp(-2L/D) \label{eq:geometric-mean-by-loss}.
    }

    As we want a bound that merely depends on the loss, we upper bound $\kappa$ using a function of the loss:
    \eql{
        \max_{\substack{\lambda_1, \dots, \lambda_D \\ \sum_i \lambda_i = D \\ \prod_i \lambda_i^{1/D} = g}} \kappa = \frac{1 + \sqrt{1 - g^D}}{1 - \sqrt{1 - g^D}}.
    }
    Then,
    \eql{
        \Ll - \Ll' \leq \frac{2}{(D + 2)} \frac{1 + \sqrt{1 - g^D}}{1 - \sqrt{1 - g^D}}^2 (1 - g).
    }

    By assumption, we have the limit $\Ll \ll 0$, which corresponds to $g \to 1$.
    We find:
    \eql{
        \Ll' \geq \left(1 - \frac{2}{D+2}\right) \Ll - |\Oo(\Ll)|
    }

\end{proof}

\subsection{Proof of \cref{thm:end-to-end-scaling-coupling}}
\label{app:end-to-end-scaling-coupling}

\begin{proof}
    The proof follows by replacing the number of parameters per layer in \cref{app:proof-end-to-end-scaling} by $(D/2)^2 + D/2$ instead of $D$. We find at least $2 > D(D+1)/(2((D/2)^2 + D/2)) = 2(D+1)/(D+2) > 1$ layers to represent $\Sigma$.
\end{proof}

\subsection{Proof of \cref{thm:iterative-convergence-rate-coupling}}
\label{app:iterative-convergence-rate-coupling}

We use Theorem 4 in \cite{draxler_whitening_2022}. Like \cref{sec:convergence-guarantees}, it concerns centers data:

\begin{assumption}
    \label{as:centered}
    The data $p(x)$ is centered: $\EE_{x \sim p(x)}[x] = 0$.
\end{assumption}

\begin{theorem}
    \label{thm:iterative-convergence-rate-coupling-original}
    Given $D$-dimensional data fulfilling \cref{as:centered,as:normalized-covariance} with covariance $\Sigma$.
    Then, after $L$ coupling blocks, the expected loss is smaller than:
    \eql{
        \EE_{Q_1, \dots, Q_L \in O(D)}[\Ss(\Sigma_L)] \leq \gamma\left(\Ss(\Sigma)\right)^L \Ss(\Sigma),
    }
    where the convergence rate depends on the non-Standardness before training:
    \eql{
        \label{eq:convergence-rate-coupling-original}
        \gamma(\Ss) = 1 + \tfrac{1}{4 \Ss/D} \log\left(1 - \frac{D^2}{(D-1)(D+2)} \frac{1 - \sqrt{1 - g(\Ss)^D}}{1 + \sqrt{1 - g(\Ss)^D}} \left(1 - g(\Ss) \right) \right) < 1.
    }
\end{theorem}

Here, the non-Standardness $\Ss$ is the loss $\Ll$ in the Gaussian case, see \cref{eq:loss-gaussian}. We write $\Ll$ in the remainder of this section:

\begin{proof}
    We start from \cref{thm:iterative-convergence-rate-coupling-original}. We first take the limit of $\gamma(\Ll)$ for $\Ll \ll 1$ and then $D \gg 1$. We use the computer algebra system \texttt{sympy} to take the limits:
    \eql{
        \gamma(\Ll) \xrightarrow{\Ll \to 0} \frac{D (D + 2) - 4}{2(D-1)(D+2)} + \Oo(\Ll) \xrightarrow{D \to \infty} \frac12 + \Oo(\Ll) + \Oo(D^{-1}).
    }
\end{proof}

To further justify the usage of
\eql{
    \gamma(\Ll) \xrightarrow{\Ll \to 0, D \to \infty} \frac12,
}
note that
\eql{
    \gamma(\Ll) \xrightarrow{\Ll \to 0} \frac{D (D + 2) - 4}{2(D-1)(D+2)} \in \left[1/2, 5/9 \right] \leq 0.555\dots
}

\section{Experimental details}

\subsection{Measuring number of required layers}
\label{app:measuring-number-of-required-layers}

We proceed as follows to predict the number of required layers to reduce the loss by a factor $\Ll'/\Ll$ from training $L_\text{train}$ layers in all experiments:

\begin{enumerate}
    \item Determine the entropy $H[p(x)]$ of the data distribution $p(x)$. This is required for computing the KL divergence in \cref{eq:loss}, and restricts us to distributions $p(x)$ which we can sample from (for training) and where we can evaluate the entropy via:
    \eql{
        H[p(x)] = -\EE_{x \sim p(x)}[\log p(x)]
    }
    \item Train a fixed number of layers $L_\text{train}$ iteratively.
    \item In general, we observe the loss as a function of depth $L$ to decrease in a geometric series:
    \eql{
        \label{eq:iterative-loss-form}
        \Ll_{L_\text{train}} := \Ll_0 \gamma^{L_\text{train}}.
    }
    We determine $\gamma$ from the initial loss $\Ll_0$ and the loss after $L_\text{train}$ layers:
    \eql{
        \label{eq:convergence-rate-computation}
        \gamma := \left[\Ll_{L_\text{train}} / \Ll_0\right]^{1/L_\text{train}}
    }
    \item We then extrapolate \cref{eq:iterative-loss-form} to predict the number of layers required for an arbitrary loss ratio $\Ll' / \Ll$:
    \eql{
        \label{eq:required-layers-given-rate}
        L = \frac{\log(\Ll'/\Ll)}{\log \gamma}
    }
\end{enumerate}

We evaluate \cref{eq:required-layers-given-rate} throughout the paper for a loss ratio of $\log(\Ll'/\Ll) = 1$, that is $\Ll' = e^{-1} \Ll = 36.8\% \Ll$, but this is arbitrary as we are only interested in the scaling with dimension which is independent of $\Ll'/\Ll$.

\subsection{Gaussian data}
\label{app:gaussian-experiment}

As our dataset, we construct the same dataset of Gaussians as \citet{draxler_whitening_2022}. They construct covariances given a dimension $D$ by choosing the eigenvalues of the covariance. We then normalize for fulfilling \cref{as:normalized-covariance} and obtain:
\eql{
    \Sigma = \frac{1}{\sum_i \lambda_i} \Diag(\lambda_1, \dots, \lambda_D).
}
We consider the following eigenvalue spectra:
\begin{enumerate}
    \item Single eigenvalue varying: $\lambda_1 = \alpha, \lambda_{>1} = 1$ for $\alpha \in (\lambda_{\min}, 1/\lambda_{\min}) \backslash \{1\}$.
    \item All eigenvalues varying but one: $\lambda_1 = 1, \lambda_{>1} = \alpha$ for $\alpha \in (\lambda_{\min}, 1/\lambda_{\min}) \backslash \{1\}$.
    \item All eigenvalues varying but one (shifted): Like the previous case, but we map $\lambda_i \mapsto \lambda_i - \sum_{i=1}^D \lambda_i / D$ and exclude spectra where this produces any $\lambda_i \leq 0$.
    \item Half small, half big: $\lambda_{\leq D/2} = \alpha, \lambda_{> D/2} = 1/\alpha$ for $\alpha \in (\lambda_{\min}, 1) \backslash \{1\}$.
    \item Randomly sample $\lambda_i \sim [0, 2]$ uniformly.
    \item Randomly sample $\lambda_i \sim [\lambda_{\min}, 1/\lambda_{\min}]$ log-uniformly.
\end{enumerate}
We choose $D$ in 10 geometrically spaced values from 10 to 128. We choose $\lambda_{\min} = 10^{-3}$. The case where all eigenvalues are equal to 1, $\lambda_i = 1$, is excluded, as Gaussianization has converged at this point.

For each $\Sigma_i$ in the resulting data set, we create $N_\text{rot}$ differently rotated variants $\Sigma_i^{r} = Q_r^\top \Sigma_i Q_r$, where we choose $N_\text{rot} = 8$, whatever is larger. This corresponds to the initial unknown rotation of the data.

We then apply the analytic solution of a Gaussianization block on the covariance, given by \cref{eq:gaussianization-block-on-gauss}. Then, another $N_\text{rot}$ rotations are drawn, which rotate each resulting covariance. This procedure is repeated until the specified number of layers is reached. We use $L_\text{train} = 8 D$ layers as we expect the number of required layers to increase with $\Omega(D)$.

To evaluate \cref{thm:iterative-convergence-rate}, we compare \cref{eq:iterative-scaling} with the empirical result. To estimate the required number of layers for a fixed loss ratio, we proceed as in \cref{app:measuring-number-of-required-layers} so that we end up in the regime where $\Ll \ll 1$. For \cref{fig:gaussian-scaling-experiment}, we fit $\gamma$ from the loss ratio of the last two layers: $\gamma = \sqrt{\Ll_{L_\text{train}}/\Ll_{L_{\text{train}} - 2}}$ instead of \cref{eq:convergence-rate-computation}. The maximum loss in this case is $10^{-2}$, so we are in the regime of $\Ll \ll 1$.

\begin{figure}[ht]
    \centering
    \includegraphics[width=.33\linewidth]{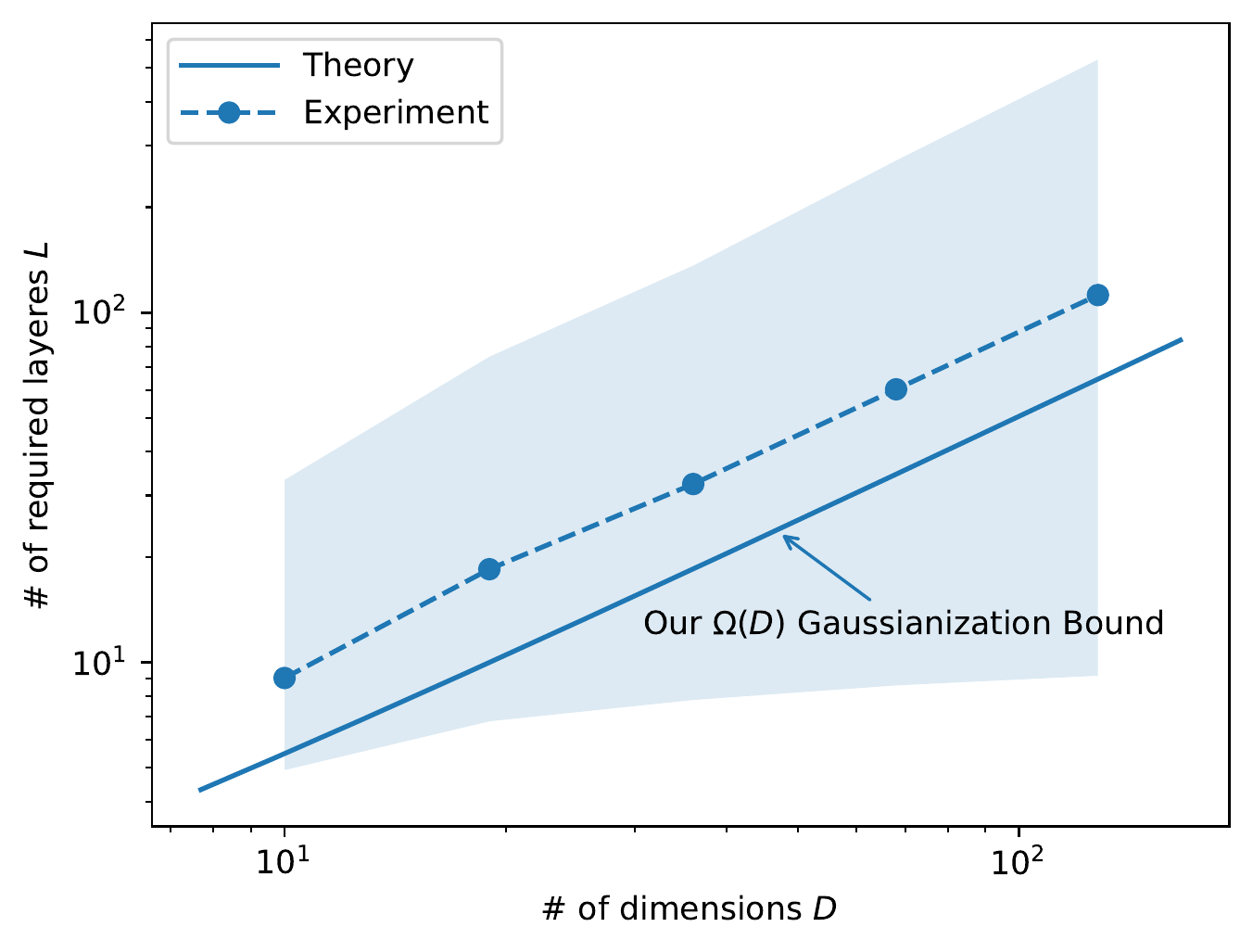}
    \includegraphics[width=.33\linewidth]{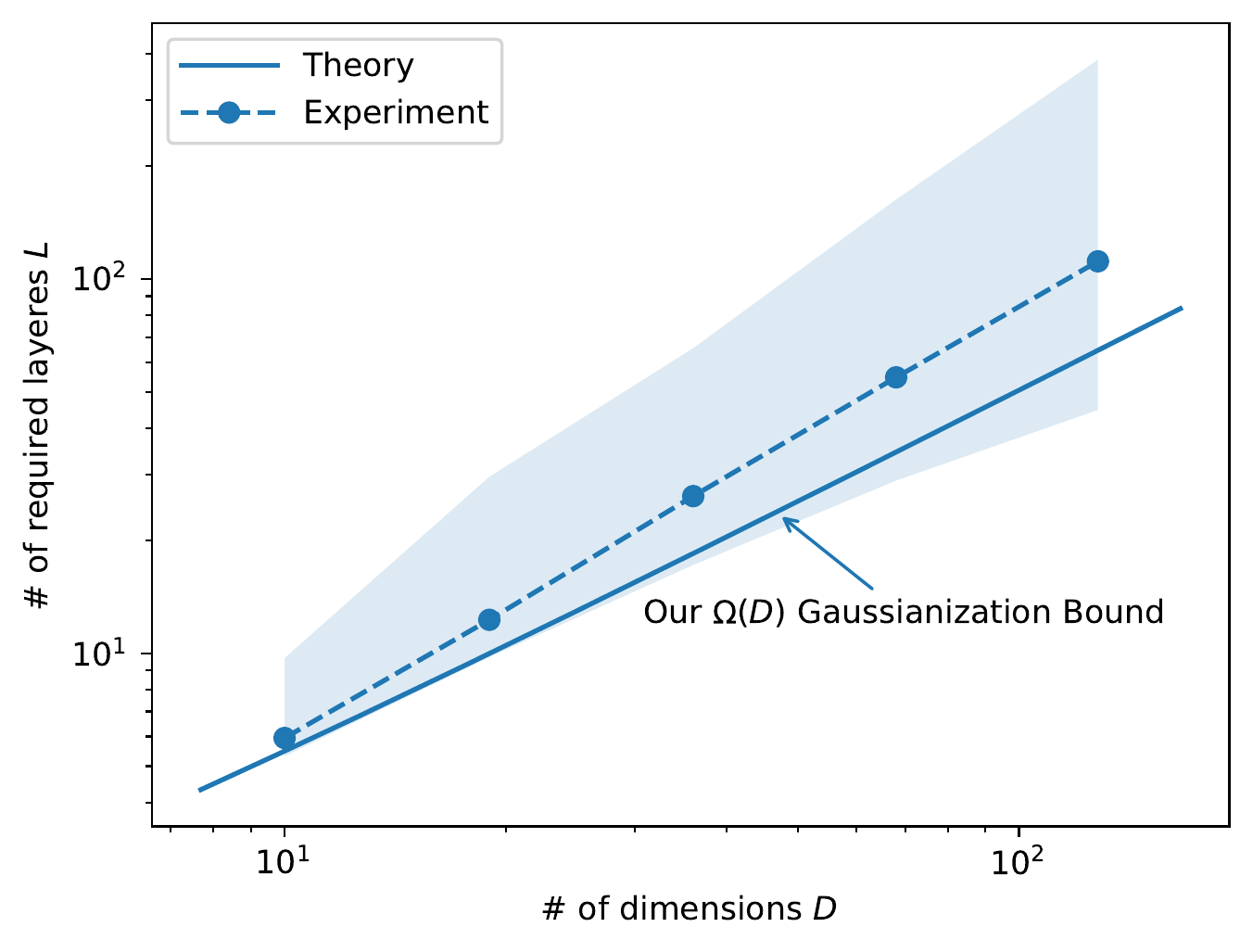}
    \includegraphics[width=.33\linewidth]{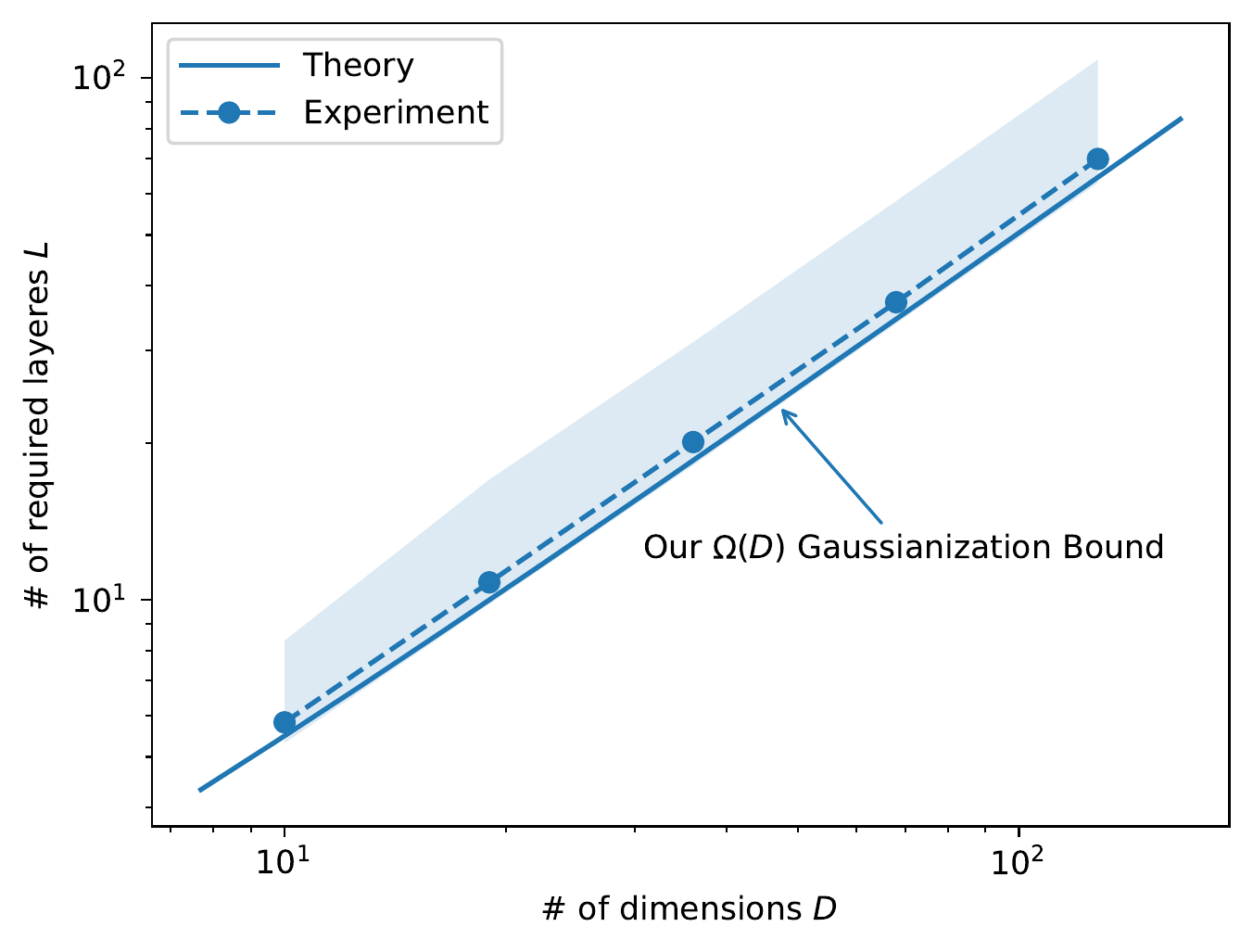}
    \caption{\textbf{For larger loss $\Ll \gg 0$, the median number of required layers over the data set is larger than predicted, and some cases scale faster.} \textit{(Left)} If we predict the number of required layers from the first $L_\text{train}=D$ layers, the majority of cases show slower convergence than predicted. Some cases show faster convergence, see \cref{fig:bound-by-case}. \textit{(Medium)} After the first $L_\text{train}=3D$ layers, most cases show the linear scaling behavior with dimension. \textit{(Right)} The bound is valid for at least 90\% of the data after $L_\text{train}=10D$ layers. All averaging is performed via the median and shaded regions cover 90\% of the cases.}
    \label{fig:bound-violations}
\end{figure}

\begin{figure}[ht]
    \centering
    \includegraphics[width=.9\linewidth]{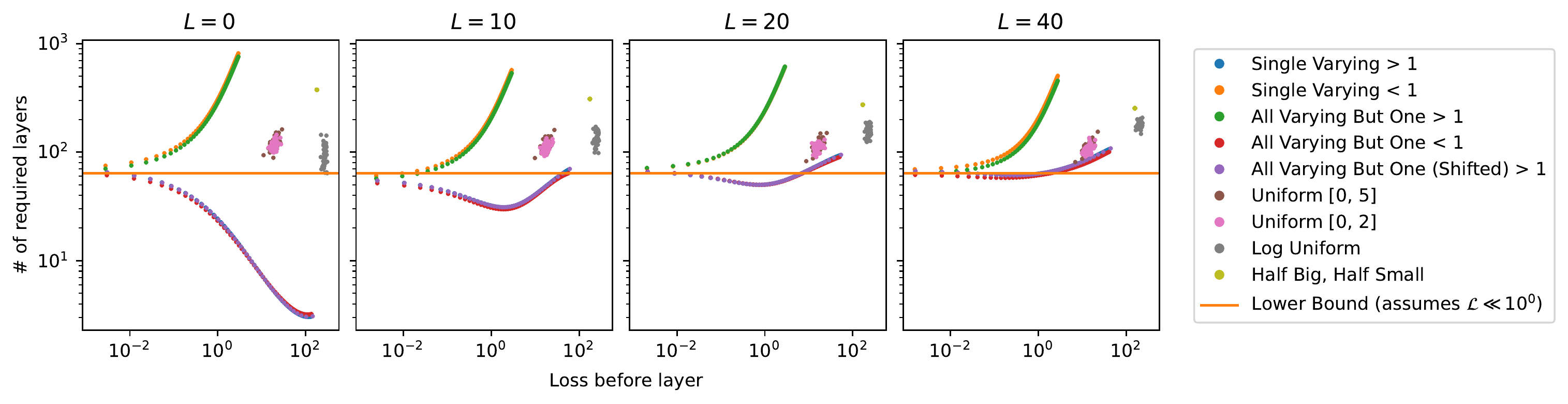}
    \caption{\textbf{Our analytic lower bound on the number of layers gives a good estimate even outside the valid regime (where $\Ll \not\ll 1$) for several considered cases.} \textit{(Left)} The cases with faster initial convergence than \cref{eq:iterative-scaling} ({\color{C1} orange}) originate from ``All eigenvalues varying but one'' and ``Single eigenvalue varying'' with one eigenvalue bigger than the others \textit{(left)}. After a finite number of layers ($L=40$ here), the number of required layers predicted at this depth is close to the number predicted by the theory in all cases. The plot considers $D=128$. Note that at this depth, the loss for the fastest configuration is still greater than 10, far from convergence. The suffixes ``$> 1$'' and ``$ < 1$'' separate $\alpha \lessgtr 1$.}
    \label{fig:bound-by-case}
\end{figure}

In \cref{fig:bound-violations}, we observe the same linear scaling behavior of required layers with dimension for $\Ll \gg 0$, i.e.~at the beginning of training. However, in this case, the bound in \cref{eq:iterative-scaling} is violated in few scenarios: They require less layers than predicted. However, after a small number of layers, also these cases fulfill \cref{eq:iterative-scaling}, see \cref{fig:bound-by-case}.

\subsection{Finding spurious dimensions in standard normal data} \label{app:spurious-rotations}

We randomly sample $N=60,000$ normal samples of dimension $D=3072$, which corresponds to the size of the CIFAR10 dataset. We optimize $w \in \RR^D, \norm{w}=1$ to minimize the 1-dimensional sample-based 2-Wasserstein distance to a predefined bimodal distribution $p_\text{adv} = \left(\Nn(-d/2, \sigma) + \Nn(d/2, \sigma)\right)/2$:
\eql{
    W_2^2 = \sum_{i=1}^N \left((w^\top x)_{\pi_w(i)} - y_i \right)^2.
}
Here, $y_i$ are sorted samples from $p_\text{adv}$. The permutation $\pi_w: [N] \to [N]$ sorts the projected values $(w^\top x)$.

We choose the spread of the bimodal distribution to be $d=2$ and the standard deviation of each mode as $\sigma=0.4$.

We optimize $w$ for 64 steps using SGD with a learning rate of 10 and momentum $.8$. After each update, we rescale $\norm{w}=1$. The final Wasserstein distance we obtain reads $0.03$, down from $0.1$ for a random $w$.

For the visualization in \cref{fig:spuriously-non-gaussian}, we project the data once with a random $w_\text{rand}^\top x$ and once with $w^\top x$ into 70 bins.

\subsection{Gaussianization implementation} \label{app:gaussianization-implementation}

Following \cref{eq:gaussianization-block}, we construct $f_{rot}$ as a random rotation $f_{rot}(x) = Qx$ where $Q \sim O(D)$ for data dimensionality $D$. For $f_{dim}$, we choose rational-quadratic splines, which allow for approximation of arbitrary functions by separating their domain into $b$ bins. Given bin edges (or knots) $x^{(k)}, y^{(k)}, x^{(k + 1)}, y^{(k + 1)}$ and derivatives $\delta^{(k)}, \delta^{(k + 1)}$ at those edges for bin $k = 1, ..., b$, they can be interpolated with a rational-quadratic polynomial as described in \cite{durkan2019neural}. Beyond the outermost bin edges, the spline is extrapolated with linear tails.

\begin{figure}[ht]
    \centering
    \includegraphics[width=0.5\linewidth]{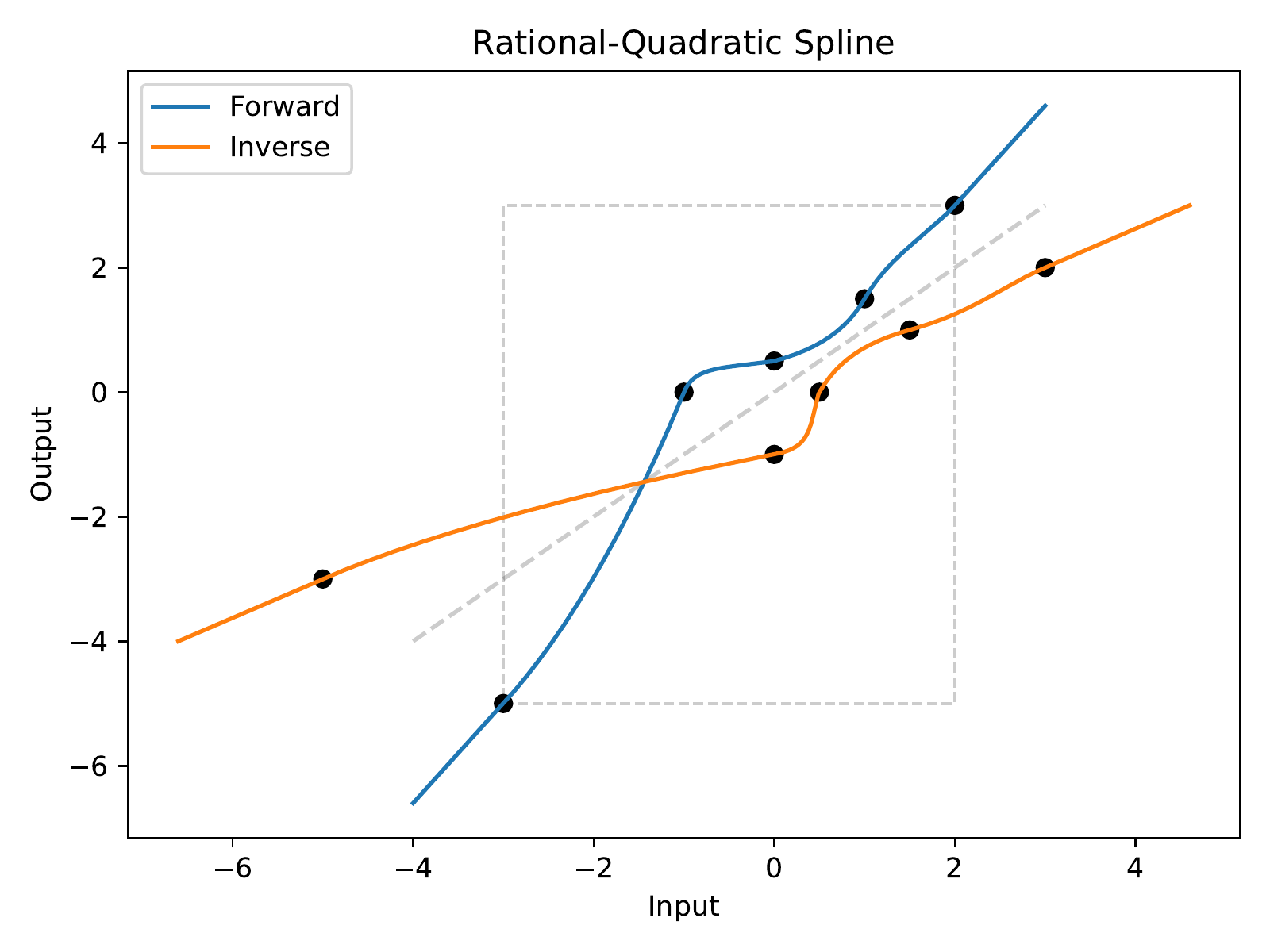}
    \caption{\textbf{Rational-Quadratic Spline.} Each pair of knots (black) is interpolated with a rational-quadratic polynomial. The outside is extrapolated with linear tails.}
    \label{fig:rq-spline}
\end{figure}

We use an implementation of RQ splines based on \cite{dai_sliced_2021}, where $\psi(x, \alpha) = (1 - \alpha) RQ(x) + \alpha x$ with a scalar regularization parameter $\alpha$. We choose $b = 128$ bins, as well as $\alpha_1 = 0.9$ for the spline and $\alpha_2 = 0.99$ for the linear extrapolation, such that
\eql{
    f_{dim, \theta_i}(x_i) = \begin{cases}
        \psi(x_i, \alpha_1) \quad x_i^{(1)} \leq x_i \leq x_i^{(b + 1)} \\
        \psi(x_i, \alpha_2) \quad \text{otherwise}
    \end{cases}.
}

The $\alpha_{1,2}$ significantly slow down training, but increase performance \cite{dai_sliced_2021}. It should not alter the scaling behavior of the number of required layers with dimension, up to a constant factor independent of dimension.

Splines are fit to the CDF of the data by evenly distributing bin knots on the quantiles of the data and applying the inverse Gaussian CDF:
\eqal{
    x_i^{(k)} &= q \left( \frac{k}{b + 2} \right), \\
    y_i^{(k)} &= G^{-1}\left( x_i^{(k)} \right),
}
where $q(p) = \sqrt{2} \operatorname{erf}^{-1}(2p - 1)$ is the quantile function and $G(x) = \frac{1}{2} \left[ 1 + \operatorname{erf} \left( \frac{x}{\sqrt{2}} \right) \right]$ is the standard normal CDF. The inner derivatives $\delta^{(k)}$ can then be estimated by finite differences, following \cite{durkan2019neural}:
\eqal{
    h_i^{(k)} &= x_i^{(k + 1)} - x_i^{(k)}, \\
    s_i^{(k)} &= \frac{y_i^{(k + 1)} - y_i^{(k)}}{x_i^{(k + 1)} - x_i^{(k)}}, \\
    \delta_i^{(k + 1)} &= \frac{s_i^{(k)} h^{(k + 1)} + s^{(k + 1)} h^{(k)}}{h^{(k + 1)} + h^{(k)}}, \quad k = 1, ..., b - 1.
}
We choose identity tails, i.e. $\delta^{(1)} = \delta^{(b + 1)} = 1$.

\subsection{Toy scaling experiment} \label{app:dependency-toy}

\paragraph{Toy data distribution}
Our goal is to create a family of distribution $p(x)$ which naturally extend over different dimensions $D$ by the continous mixture of Gaussians \cref{eq:toy-distribution-details} where the mean of each dimension is conditioned on the previous via $m_i(x_j \in A_i)$. In particular, we choose:
\eql{
    m_1 = \frac12, \quad
    m_0 = 0, \quad
    \sigma_1^2 = 0.8, \quad
    \sigma_2^2 = 0.2.
}
The $s_{ij}(D)$ are drawn randomly from $\{-1, +1\}$ for each dimension $D$ and for each seed. This is the main source of noise between runs.

\begin{figure}
    \centering
    \includegraphics[width=.5\linewidth]{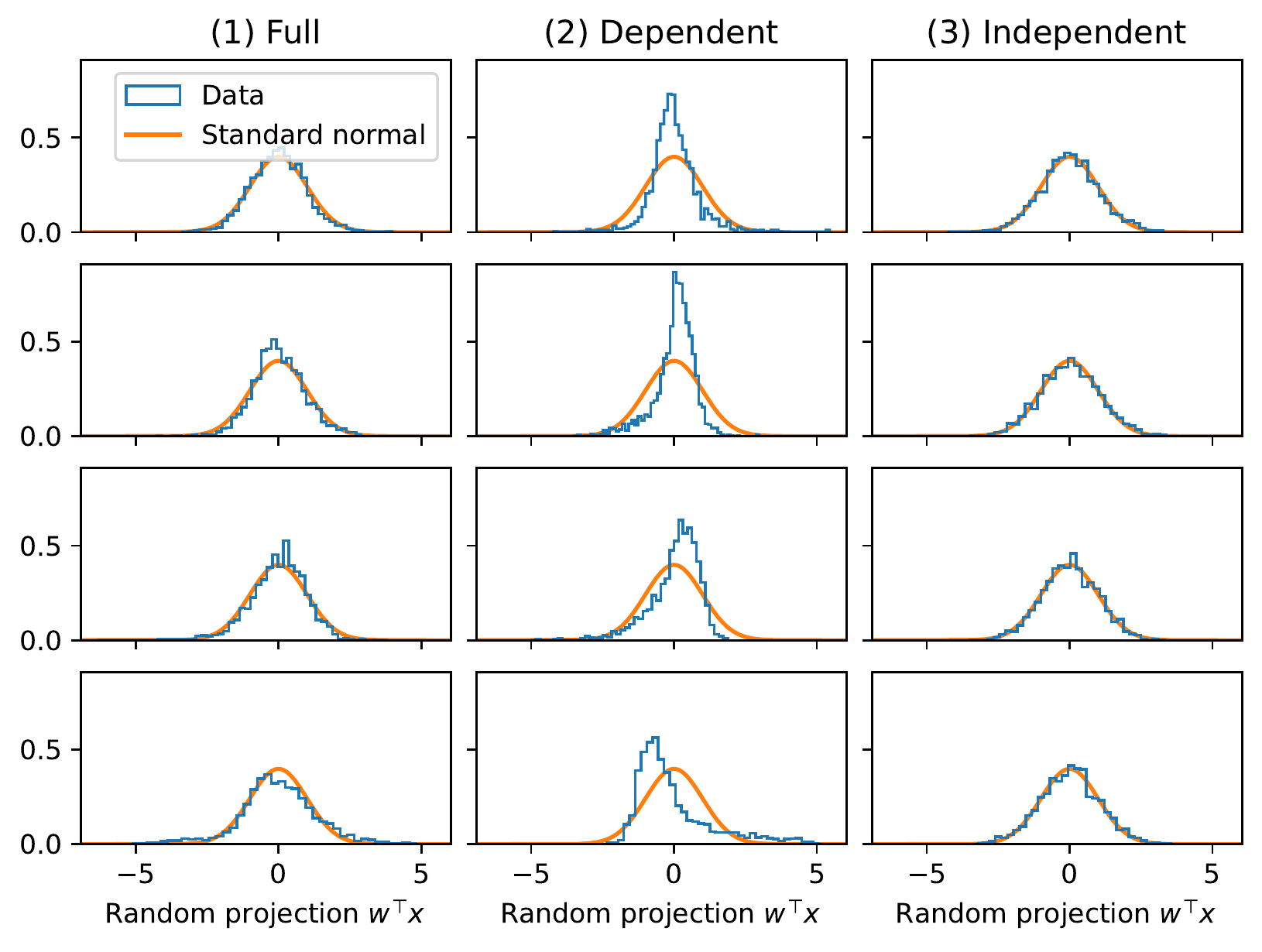}
    \caption{Random projections of data from the different versions of toy data in $D=128$. Columns show the different cases, and each row depicts a different random projection $w \in \RR^D, \norm{w} = 1$. In case (2, \textit{center}), most dimensions depend on the same dimensions, making random projections deviate strongly from a standard normal. For case (1) and (3), random projections are very close to a Gaussian, preventing a fast fit using Gaussianization. %
    }
    \label{fig:toy-random-projections}
\end{figure}

\paragraph{Training} We employ the architecture described in \cref{app:gaussianization-implementation} for $L_\text{train}=64$ layers. We use $N = 60,000$ training samples in each case and run a total of four runs per case (except for case 1, for which we average over eight runs). Before training, the data is normalized to zero mean and unit standard deviation in the original rotation.

In principle, measuring convergence rates would be more accurate by using different batch of training data for fitting each layer, as this avoids overfitting and yields slower convergence on test data. We find this not to be a problem for the considered 64 layers and high $\alpha$ (see \cref{app:gaussianization-implementation}).

\paragraph{Evaluation} We compute the number of required layers for each run via the procedure in \cref{app:measuring-number-of-required-layers}. This number is then averaged over all runs for each case and dimension.

\subsection{Multi-scale EMNIST experiment} \label{app:dependency-real}

\paragraph{Data distribution} As described in \cref{sec:dependency-real}, we make use of a normalizing flow as our data density $p(x)$.
Each flow architecture consists of a fixed normalization layer, followed by 20 affine coupling blocks. We use purely full-connected networks for odd scales, and convolutional networks for even scales. We use wavelet downsampling before the first. If the image side length scale is a power of 4, we add a second wavelet downsampling after the eigth affine coupling layer. Each convolutional subnetwork uses two hidden layers with 16 channels respectively 32 channels after the second downsampling each, and a kernel size of 3. The final 4 coupling blocks are fully connected. Each fully connected subnetwork has two hidden layers with hidden width equal to the total number of dimensions $D$. The details for each architecture are given in \cref{tab:distribution_flows}.

We train the normalizing flows for 30 epochs for $D=28 \times 28$, and 20 for the other scales using negative log-likelihood, see \cref{eq:loss}. We use Adam with a learning rate of $10^{-3}$ and a batch size of $256$.

After training, we replace the latent distribution with a normal distribution $\Nn(0, \hat\sigma^2 I_{D})$ with $\hat\sigma = 0.8$ to achieve a better sample quality. Note that this does not influence the ability to compute density estimates $p(x)$ from our model, but it does reduce the entropy of the data.

\begin{table}[ht]
    \centering
    \begin{tabular}{|c|c|c|c|}
        \hline
        Scale & Network type & \# of downsamplings  & \# of parameters \\
        \hline
		$4 = 2 \times 2$ & conv & 1 & 149k \\
		$9 = 3 \times 3$ & fc & 0 & 7k \\
		$16 = 4 \times 4$ & conv & 2 & 194k \\
		$25 = 5 \times 5$ & fc & 0 & 38k \\
		$36 = 6 \times 6$ & conv & 1 & 164k \\
		$49 = 7 \times 7$ & fc & 0 & 134k \\
		$64 = 8 \times 8$ & conv & 2 & 235k \\
		$100 = 10 \times 10$ & conv & 1 & 254k \\
		$144 = 12 \times 12$ & conv & 2 & 406k \\
		$196 = 14 \times 14$ & conv & 1 & 544k \\
		$256 = 16 \times 16$ & conv & 2 & 861k \\
		$324 = 18 \times 18$ & conv & 1 & 1M \\
		$400 = 20 \times 20$ & conv & 2 & 2M \\
		$484 = 22 \times 22$ & conv & 1 & 3M \\
		$576 = 24 \times 24$ & conv & 2 & 4M \\
		$676 = 26 \times 26$ & conv & 1 & 5M \\
		$784 = 28 \times 28$ & conv & 2 & 6M \\
        \hline
    \end{tabular}
    \caption{Normalizing Flow architecture as a function of image size. A purely fully-connected network is labeled by ``fc'', ``conv'' networks are partially convolutional.}
    \label{tab:distribution_flows}
\end{table}

\begin{figure}
    \centering
    \includegraphics[width=0.5\linewidth]{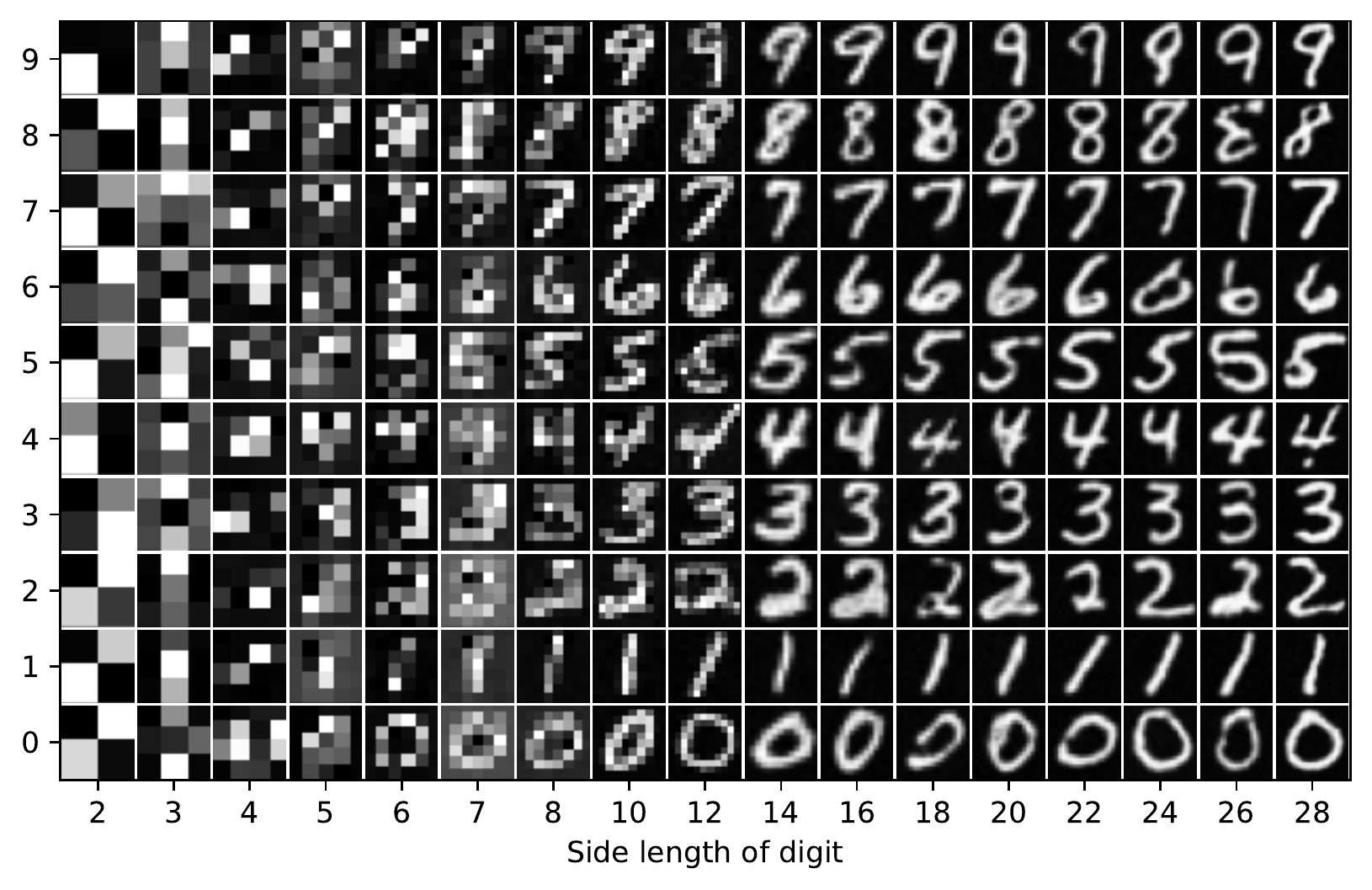}
    \caption{Samples from our down-scaled EMNIST normalized models which we use as ground truth distributions $p(x)$. Normalizing Flows are sampled with reduced temperature $0.8$.}
    \label{fig:enter-label}
\end{figure}

\paragraph{Traininig} Like in the toy experiment, we use the implementation from \cref{app:gaussianization-implementation}. We again choose $N=60,000$ training samples without resampling. We average each case over 10 runs.

\paragraph{Evaluation} We use the same procedure as in \cref{app:dependency-toy}.

\subsection{Compute and libraries}

Experiments were performed on three workstations, each with a single high-end consumer GPU and CPU each.
We build our code upon the following python libraries: PyTorch \citep{paszke2019pytorch}, PyTorch Lightning \citep{Falcon_PyTorch_Lightning_2019}, Lightning Trainable \citep{trainable2023}, Tensorflow \citep{tensorflow2015-whitepaper} for FID score evaluation, Numpy \citep{harris2020array}, Matplotlib \citep{Hunter:2007} for plotting and Pandas \citep{mckinney-proc-scipy-2010,reback2020pandas} for data evaluation.

\end{document}